\documentclass{article} 
\usepackage{iclr2026_conference,times}

\usepackage{amsmath,amsfonts,bm}

\def\eqref#1{equation~\ref{#1}}

\def\1{\bm{1}}

\DeclareMathAlphabet{\mathsfit}{\encodingdefault}{\sfdefault}{m}{sl}
\SetMathAlphabet{\mathsfit}{bold}{\encodingdefault}{\sfdefault}{bx}{n}

\usepackage[hidelinks]{hyperref}
\usepackage{url}
\usepackage{booktabs}       
\usepackage{amsfonts}       
\usepackage{nicefrac}       
\usepackage{microtype}      
\usepackage{xcolor}         

\usepackage{colortbl}
\usepackage{graphicx}
\usepackage{cleveref}
\usepackage{ragged2e}
\usepackage{multirow}
\usepackage{tcolorbox}
\usepackage{floatrow}
\usepackage{array}
\usepackage{makecell}
\usepackage{color,soul}
\usepackage{wrapfig}
\usepackage{float}
\usepackage{caption}

\definecolor{myblue}{HTML}{336de9} 
\hypersetup{colorlinks=true, citecolor=myblue, linkcolor=myblue, urlcolor=myblue} 
\floatstyle{plaintop}
\restylefloat{table}

\DeclareFloatFont{tiny}{\tiny}
\renewcommand{\arraystretch}{1.5}

\definecolor{green2}{HTML}{E8EEE7}
\definecolor{green3}{HTML}{C1DBBD}
\definecolor{green4}{HTML}{AFD1A9}
\definecolor{green5}{HTML}{9CC695}

\definecolor{red5}{HTML}{FDDA7A}
\definecolor{red4}{HTML}{FEE398}
\definecolor{red3}{HTML}{FEEBB5}
\definecolor{red2}{HTML}{FEF1CD}

\newcommand{\getcellcolor}[1]{%
    \ifdim #1 pt > 4 pt \cellcolor{green5}%
    \else\ifdim #1 pt > 2 pt \cellcolor{green4}%
    \else\ifdim #1 pt > 1 pt \cellcolor{green3}%
    \else\ifdim #1 pt > 0 pt \cellcolor{green2}%
    \else\ifdim #1 pt = 0 pt \cellcolor{white}%
    \else\ifdim #1 pt > -1 pt \cellcolor{red2}%
    \else\ifdim #1 pt > -2 pt \cellcolor{red3}%
    \else\ifdim #1 pt > -4 pt \cellcolor{red4}%
    \else \cellcolor{red5}%
    \fi\fi\fi\fi\fi\fi\fi\fi
}
\newcommand{\highlightcell}[2]{%
    \getcellcolor{#2} #1
}

\newboolean{isBlue}
\setboolean{isBlue}{False} 
\newcommand{\pr}[1]{%
    \ifthenelse{\boolean{isBlue}}%
    {\textcolor{teal}{#1}}
    {#1}
}

\title{Reasoning on a Spectrum: Aligning LLMs to System 1 and System 2 Thinking}

\author{ Alireza S. Ziabari\thanks{Equal contribution.}
\And Nona Ghazizadeh\footnotemark[1] 
\And Zhivar Sourati  \AND   Farzan Karimi-Malekabadi
\And Payam Piray \And Morteza Dehghani \AND \normalfont   University of Southern California \\
}

\iclrfinalcopy 
\begin{document}

\maketitle

\begin{abstract}
Large Language Models (LLMs) exhibit impressive reasoning abilities, yet their reliance on structured step-by-step processing reveals a critical limitation. In contrast, human cognition fluidly adapts between intuitive, heuristic (System 1) and analytical, deliberative (System 2) reasoning depending on the context. This difference between human cognitive flexibility and LLMs' reliance on a single reasoning style raises a critical question: while human fast heuristic reasoning evolved for its efficiency and adaptability, is a uniform reasoning approach truly optimal for LLMs, or does its inflexibility make them brittle and unreliable when faced with tasks demanding more agile, intuitive responses? To answer these questions, we explicitly align LLMs to these reasoning styles by curating a dataset with valid System 1 and System 2 answers, and evaluate their performance across reasoning benchmarks. Our results reveal an accuracy-efficiency trade-off: System 2-aligned models excel in arithmetic and symbolic reasoning, while System 1-aligned models perform better in commonsense reasoning tasks. To analyze the reasoning spectrum, we interpolated between the two extremes \pr{by varying the proportion of alignment data, which resulted in} a monotonic change in accuracy. A mechanistic analysis of model responses shows that System 1 models employ more definitive outputs, whereas System 2 models demonstrate greater uncertainty. \pr{Building on these findings, we further combine System 1- and System 2-aligned models based on the entropy of their generations, without additional training, and obtain a dynamic model that outperforms across nearly all benchmarks.} This work challenges the assumption that step-by-step reasoning is always optimal and highlights the need for adapting reasoning strategies based on task demands. 
\footnote{Our data and code are available at \url{https://github.com/AlirezaZiabari/System_1_System_2_Alignment}}
\end{abstract}

\section{Introduction}
LLMs have demonstrated remarkable reasoning capabilities, often achieving near-human or even superhuman performance \citep{huang-chang-2023-towards}. These advances have largely been driven by techniques that simulate step-by-step, deliberative reasoning, such as Chain-of-Thought (CoT) prompting and inference-time interventions \citep{wei2022chain, wang2022self}. Given their success, such methods are increasingly integrated into LLM training \citep{chung2024scaling}, reinforcing explicit, structured reasoning regardless of the task necessity. However, the increasing focus on step-by-step reasoning has revealed limitations such as brittle generalization, particularly in tasks requiring nuanced judgment \citep{deletang2023neural}, logical consistency \citep{jiang-etal-2024-peek}, or adaptability to uncertainty \citep{mirzadeh2024gsm}. Similarly, recent analyses frame this issue as ``overthinking'' \citep{cuadron2025danger}; \citet{chen2024not} demonstrate that excessive deliberation can hamper decision-making. This problem appears in LLMs' responses to simple factual queries, where they often generate unnecessarily long explanations instead of direct responses \citep{wang-etal-2023-plan}.

This focus on explicit, structured reasoning highlights a key difference between LLMs and human cognition: while LLMs are being pushed towards a single mode of processing, human reasoning is far more nuanced. Rather than a monolithic process, human reasoning emerges from a repertoire of cognitive tools evolved to tackle a \emph{spectrum} of computational problems. This spectrum encompasses both automatic and reflective processes, a key insight recognized across diverse fields from economics to psychology and neuroscience \citep{daw2005uncertainty, dolan2013goals, balleine1998goal}. On one end lie computationally \emph{light} problems demanding rapid, intuitive judgments with confidence (e.g., instinctively dodging a speeding car), handled by the reflexive ``System 1 ($\mathcal{S}1$).''  On the other end are \emph{heavy} problems requiring deliberate, step-by-step analysis with prospection, managed by the reflective ``System 2 ($\mathcal{S}2$)'' \citep{kahneman2011thinking, stanovich2000advancing}. This dual-process system allows us to dynamically shift between modes depending on the task, balancing speed and accuracy \citep{evans2013dual}. Extensive work in neuroscience over the past two decades links the dual-process framework and human decision strategies, which depicts decision-making on a spectrum between a fast but reflexive habitual decision strategy \pr{\citep{daw2011model, gillan2016characterizing, miller2017dorsal}} and a reflective goal-directed strategy \citep{daw2005uncertainty, dolan2013goals}. Experimental work in neuroscience is built on the relative advantages of these strategies, the separate but overlapping neural structures supporting them, and the circumstances under which each system is deployed in the brain \citep{schad2020dissociating, piray2021linear}. Given the evolutionary advantage of switching between fast and slow thinking to balance speed, efficiency, and accuracy, exploring LLMs through the lens of dual-process theory offers a powerful way to address their limitations.

While recent studies explore whether LLMs exhibit $\mathcal{S}1$ and $\mathcal{S}2$ behaviors \citep{hagendorff2023human, pan2024dynathink} or propose hybrid models \citep{yang2024llm2, deng2024cognidual}, most prior work implicitly assumes that structured, deliberative reasoning is universally superior. Even research suggesting LLMs' capacity for both reasoning modes \citep{wang2024chain} largely overlooks the crucial question of when each mode is indeed advantageous. The assumption that a single ``best'' reasoning strategy can apply across all contexts is a fundamental simplification that limits current approaches in LLM development. This assumption prevents LLMs from achieving human-like cognitive flexibility, hindering their ability to adapt their reasoning processes to diverse situations.

\pr{To address this gap, we design an experimental setup where both thinking styles can produce valid responses but follow distinct paths, one leveraging intuitive heuristics, and the other prioritizing deliberate, step-by-step reasoning. To implement this setup, we first curate a dataset of 2,000 reasoning questions where each problem has both $\mathcal{S}1$ and $\mathcal{S}2$ responses, grounded in ten well-studied cognitive heuristics \citep{tversky1974judgment}. Next, we explicitly align LLMs with either $\mathcal{S}1$ or $\mathcal{S}2$ responses and systematically assess them across diverse reasoning benchmarks. Our findings mirror the well-known accuracy–efficiency trade-off in human cognition \citep{keramati2011speed, mattar2018prioritized}: $\mathcal{S}2$–aligned models excel in arithmetic and symbolic reasoning, demonstrating superior multi-step inference but producing longer, token-intensive outputs, while $\mathcal{S}1$–aligned models generate succinct answers and perform better on commonsense reasoning tasks where heuristic shortcuts are effective. 
Beyond this trade-off, we also show that $\mathcal{S}1$ models are more confident and decisive, whereas $\mathcal{S}2$ models express greater uncertainty and hedge more, mirroring patterns observed in neuroscience \citep{daw2005uncertainty}.
Then, to further examine this spectrum, we interpolated between the two extremes by varying the proportion of alignment data, which yielded a monotonic change in accuracy. Finally, we propose a training-free dynamic model that adaptively chooses between $\mathcal{S}1$ and $\mathcal{S}2$ reasoning based on output entropy signals. By framing LLM reasoning as a structured and adaptable process, this work highlights the importance of selecting the right reasoning strategy for a given task and sets the stage for more flexible, efficient, and robust reasoning systems.}

\section{Related Work}
\paragraph{Reasoning in LLMs.}
Extensive research highlights both the strengths and weaknesses of LLM reasoning \citep{huang2022towards,mondorf2024beyond,valmeekam2022large,parmar2024logicbench,sourati2024arn, shojaee2025illusion}. Recent efforts to enhance these abilities have largely focused on prompting \citep{brown2020language}, from zero-shot prompting with explicit instructions \citep{kojima2022large, wang-etal-2023-plan, zhou2024self} to few-shot prompting with step-by-step examples \citep{wei2022chain}. \citet{wang2024chain} further show that CoT reasoning can be elicited from pre-trained LLMs by output decoding without a CoT prompt. Self-consistency decoding \citep{wang2022self} improves robustness through diverse reasoning paths, aligning with $\mathcal{S}2$ reasoning. Tree of Thought \citep{yao2024tree} generalizes CoT, enabling LMs to explore multiple reasoning paths, self-evaluate, and look ahead or back to make global decisions.
LLM reasoning can also be improved via CoT instruction tuning \citep{chung2024scaling, huang2022large} or distillation \citep{magister2022teaching}, enabling models to internalize step-by-step reasoning and surpass prompting techniques. 
Concurrent studies have identified an ``overthinking'' phenomenon in LLMs, where models generate excessively detailed or unnecessary reasoning steps \citep{chen2024not, cuadron2025danger}.

\paragraph{Dual-Process Theory in LLMs.}
Dual-process theory offers a powerful framework for understanding human reasoning, though its use in NLP is still relatively underexplored. Existing research broadly falls into two main categories: 
First, researchers have investigated whether LLMs exhibit reasoning behaviors aligned with $\mathcal{S}1$ and $\mathcal{S}2$, particularly in terms of cognitive human-like errors and biases \citep{pan2024dynathink, echterhoff2024cognitive, zeng2024mr}. Specifically, \citet{hagendorff2023human} examine cognitive heuristics in LLMs, showing that newer models exhibit fewer errors characterize with $\mathcal{S}1$ thinking. \citet{booch2021thinking} discuss fundamental questions regarding the role of dual-process theory in ML but leave practical implementation as an open problem. 
Second, several studies have integrated dual-process-inspired reasoning into LLMs. Some works combine intuitive (fast) and deliberate (slow) components to improve reasoning and planning \citep{he-etal-2024-planning, liu2022neural, hua-zhang-2022-system, pan2024dynathink, su2025dualformer, saha2025system}, while others optimize efficiency by distilling $\mathcal{S}2$ insights into $\mathcal{S}1$ models \citep{yang2024llm2, deng2024cognidual, yu2024distilling}. Research has also leveraged $\mathcal{S}2$ reasoning to mitigate biases associated with $\mathcal{S}1$ heuristics to improve fairness and robustness \citep{furniturewala2024thinking, kamruzzaman2024prompting, weston2023system}.
While most studies frame $\mathcal{S}2$ as superior and portray $\mathcal{S}1$ as erroneous despite its role in efficient reasoning, we instead investigate the implicit effects of aligning LLMs to either system. By analyzing how these heuristics shape general reasoning, we address a gap in the literature and offer new insights into broader cognitive behaviors of LLMs.

\section{Method}
\subsection{Aligning LLMs to System 1 \&  System 2 Thinking}\label{subsec:aligning-to-system1-2-thinking}
We formalize fast and slow thinking as an alignment problem using a curated dataset where each reasoning question is paired with both $\mathcal{S}1$ (intuitive) and $\mathcal{S}2$ (analytical) responses (see \Cref{sec:dataset}). We align LLMs to either reasoning style via a preference-based training approach: for $\mathcal{S}1$ alignment, the intuitive response is designated as the preferred (winner) and the analytical response as the non-preferred (loser); for $\mathcal{S}2$ alignment, this preference is reversed, treating the analytical response as the winner and the intuitive response as the loser. 

This approach is effective for two reasons: 
First, previous research has shown that prompt engineering can guide LLMs toward $\mathcal{S}2$ reasoning \citep{wei2022emergent} or $\mathcal{S}1$ reasoning \citep{zhou-etal-2024-llms}, suggesting that LLMs already have both reasoning abilities. Therefore, instead of creating new reasoning pathways, we guide the model to distinguish between intuitive and analytical reasoning without altering its underlying knowledge. 
Second, our aim is not to introduce new knowledge or instructions but rather to shape the model’s reasoning process based on existing capabilities. 

\begin{figure}[t]
\renewcommand{\arraystretch}{0.9}
\centering
  \includegraphics[width=\textwidth]{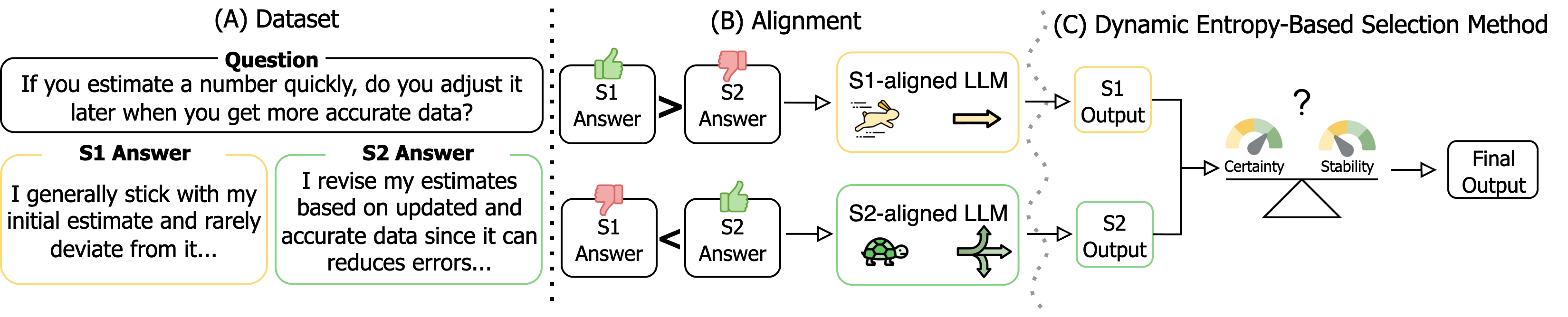}
  \caption{(A) Sample of dataset with System 1 and System 2 answers. (B) Overview of our alignment approach with fast and slow thinking. (C) Overview of our dynamic entropy based selection method.}
  \label{fig:overview}
\end{figure}

\subsection{Entropy-Based Arbitration Between Reasoning Styles} \label{sec:dynamicmethod}
\pr{
To create a dynamic model, we propose a training-free approach that arbitrates between $\mathcal{S}1$– and $\mathcal{S}2$-aligned models dynamically. The method adaptively selects the reasoning style best suited to a given query using entropy-based signals. To quantify LLM uncertainty, we compute token-level entropy for each generated sequence of tokens $T = (t_1, \ldots, t_n)$ over vocabulary $V$:}
\begin{equation}
    H_i = -\sum_{v \in V} P(v | t_{<i}, x) \log P(v | t_{<i}, x),
\end{equation}
\pr{where $P(v | t_{<i}, x)$ is the probability of token $v$ given the input $x$ and preceding tokens $t_{<i}$. From these token-level entropies, we calculate the average sequence entropy $\bar{H}$ and its variance $\sigma^2$:}
\begin{equation} \label{eq:ent}
    \bar{H} = \frac{1}{n} \sum_{i=1}^{n} H_i , \quad
    \sigma^2 = \frac{1}{n} \sum_{t=1}^{n} (H_i - \bar{H})^2.
\end{equation}
\pr{
$\bar{H}$ captures the overall uncertainty of the model’s predictions, while $\sigma^2$ reflects the instability of its reasoning process. ``Stable and confident'' predictions correspond to low values of both, ``cautious but consistent'' predictions arise from high $\bar{H}$ with low $\sigma^2$, and ``instability'' is signaled by high $\sigma^2$ regardless of $\bar{H}$. To enable comparison between $\mathcal{S}1$ and $\mathcal{S}2$ models, we denote their entropy statistics as $(\bar{H}_1, \sigma^2_1)$ and  $(\bar{H}_2, \sigma^2_2)$, and normalize them via total sum scaling across the two systems, yielding $(\hat{H}_1, \hat{\sigma}^2_1)$ and  $(\hat{H}_2, \hat{\sigma}^2_2)$. We then define the reliability $R_i$ for each model as a combined score:}

\begin{equation}\label{eq:combined_score}
    R_i = w \times \hat{H}_i + (1-w) \times \hat{\sigma}^2_i, \quad 0\leq w\leq1.
\end{equation}
\pr{
For each question, the system with the lower score is selected. Recent works on reasoning stability \citep{you2025probabilistic, he2025can, ling2025wakenllm} suggest penalizing instability more heavily than caution ($0\leq w<\frac{1}{2}$). This scheme prioritizes ``stable and confident'' reasoning, accepts ``cautious but consistent'' reasoning, and penalizes ``unstable'' reasoning. In this way, the dynamic model outputs the most reliable answer between either $\mathcal{S}1$ or $\mathcal{S}2$ based on entropy signals without additional training.
}

\subsection{Dataset of System 1 \& System 2 Thinking} \label{sec:dataset}

Our curated dataset consists of 2,000 questions\pr{, each paired with two responses that capture distinct reasoning styles in English:}
one intuitive and rapid, reflecting cognitive shortcuts ($\mathcal{S}1$), and the other deliberate and analytical ($\mathcal{S}2$). This dual structure provides a controlled setting to examine the mechanisms underlying $\mathcal{S}1$ and $\mathcal{S}2$ reasoning \citep{kahneman2011thinking, stanovich2000advancing, evans2013dual}. The dataset was constructed in three phases:

\paragraph{Generation.}
To construct our dataset, we adopted a human-in-the-loop pipeline with GPT-4o \citep{hurst2024gpt} to scale high-quality reasoning examples. In line with recent work on dataset creation using LLMs \citep{xu2023wizardlm, wang2022self}, we used a one-shot prompting setup, where each generation is guided by a seed example grounded in a cognitive heuristic, providing a practical foundation for distinguishing $\mathcal{S}1$ from $\mathcal{S}2$ reasoning \citep{kahneman2011thinking}. These seed examples, authored by domain experts (i.e., cognitive scientists; see \Cref{appendix:expertsdetails}), cover 10 well-known heuristics from \citet{kahneman2011thinking} (\Cref{appendix:cognitivebiases}). For each heuristic, experts provided a reasoning question with both a $\mathcal{S}1$ (heuristic) and $\mathcal{S}2$ (deliberative) response. During expansion, the prompt included the heuristic definition, descriptions of both systems approaches, and the expert-written example, enabling the model to generate new reasoning items aligned with distinct cognitive patterns.
\pr{Early experiments showed that outcome-focused examples did not meaningfully guide model behavior. Thus, rather than mimicking naturalistic human responses, we designed process-oriented examples that explicitly articulate $\mathcal{S}1$ and $\mathcal{S}2$ reasoning. This helped models internalize distinct reasoning strategies beyond surface-level responses, as further supported in \Cref{sec:results,res:uncertainty,appendix:failure_patterns}
}.
Prompt details and expert-authored examples are in \Cref{appendix:dataexpertsamples,appendix:dataexpansionprompt}.

\paragraph{Refinement.} 
As a byproduct of the data generation process, $\mathcal{S}2$ outputs were significantly longer and more detailed, reflecting step-by-step reasoning, while $\mathcal{S}1$ outputs were shorter and more direct (\textit{Welch}’s test: $t(2090.1) = -184.74$, $p <.001$, $d = -5.84$). 
Prior work demonstrates that alignment methods can rely on superficial cues, such as output length, favoring longer responses even without reasoning advantages \citep{Singhal2023ALW}. To prevent this bias, we use zero-shot prompting with GPT-4o to match the lengths of our $\mathcal{S}1$ and $\mathcal{S}2$ outputs while preserving content. Adjustments were applied only for significant length disparity. Details on the prompt and the length disparity threshold are in \Cref{appendix:length}. By reducing the length disparity, we minimized any preference for $\mathcal{S}2$ outputs arising from their longer responses. After adjustment, $\mathcal{S}1$ outputs averaged 82.19 tokens, while $\mathcal{S}2$ outputs averaged 83.93 tokens. A two one-sided t-test (TOST) confirmed the equivalence of post-adjustment lengths across various token counts as equivalence margins (see~\Cref{appendix:length tost}),
indicating that the adjustment effectively eliminated significant length differences between the response types.

\paragraph{Verification.} Prior works show that high-quality, expert-supervised datasets of this scale are common and effective for LLM fine-tuning  \citep{xiao-etal-2024-healme, dumpala2024sugarcrepe++, li2024language}. Following this precedent to ensure data quality, we had our expert cognitive scientists conform all generated data to formal definitions of $\mathcal{S}1$ and $\mathcal{S}2$ thinking, and ensured that the dataset covers the intended set of cognitive heuristics across varied subjects. In this process, experts manually revised approximately 20\% of the responses. We further verified the breadth of topic coverage via topic modeling; for more details and a sample of the curated dataset, see \Cref{appendix:dataset-sample,appendix:topicmodeling}.

\section{Experiments Setup}\label{experiment_setup}
\paragraph{Alignment Algorithm.}
To implement the alignment strategy for $\mathcal{S}1$ and $\mathcal{S}2$ reasoning, we utilize two offline preference optimization methods, namely, Direct Preference Optimization \citep[DPO;][]{rafailov2024direct} and Simple Preference Optimization \citep[SimPO;][]{meng2024simpo}, \pr{for two reasons:} (i) their offline formulation removes the costly on-policy sampling loop, yielding a simpler and more compute-efficient training pipeline, and (ii) our hand-crafted preference pairs capture fine-grained relational signals that would likely be blurred by online-generated pairs (more details in \Cref{appendix:alignmentalgorithm}). 

\paragraph{Benchmarks.}
We evaluate our models on 13 reasoning benchmarks across three different categories: (1) arithmetic reasoning: MultiArith \citep{roy-roth-2015-solving}, GSM8K \citep{cobbe2021trainingverifierssolvemath}, AddSub \citep{hosseini-etal-2014-learning}, AQUA-RAT \citep{ling-etal-2017-program}, SingleEq \citep{10.1162/tacl_a_00160}, and SVAMP \citep{patel-etal-2021-nlp}; (2) commonsense reasoning: CSQA \citep{talmor-etal-2019-commonsenseqa}, StrategyQA \citep{10.1162/tacl_a_00370}, PIQA \citep{bisk2020piqa}, SIQA \citep{sap2019socialiqa}, and COM2SENSE \citep{singh-etal-2021-com2sense}; (3) symbolic reasoning: Last Letter Concatenation and Coin Flip \cite{wei2022chain}.
Following \citet{kong2024betterzeroshotreasoningroleplay}, our evaluation follows a two-stage process. In the first stage, we present benchmark questions to model and record their responses. In the second stage, we prompt the model with the original question, its initial response, and benchmark-specific instructions to ensure the output is formatted as required. See \Cref{appendix:benchmarkinstruction,appendix:benchmarkdetails} for benchmark details and instructions.

\paragraph{Implementation Details.}
We use Llama-3-8B-Instruct \citep{llama3modelcard} and Mistral-7B-Instruct-v0.1 \citep{jiang2023mistral7b} as SFT models for alignment. Following \citet{kojima2023largelanguagemodelszeroshot}, we compare the performance of these aligned models against their instruction-tuned counterparts under zero-shot and zero-shot CoT prompting (details in \Cref{appendix:impdet}). To analyze the model's behavior along the $\mathcal{S}1$ to $\mathcal{S}2$ reasoning spectrum, we train seven intermediate models, where the winner responses are mixed at predefined ratios between $\mathcal{S}1$ and $\mathcal{S}2$. This structured interpolation allows us to systematically assess whether the transition between reasoning styles is discrete or gradual.

\section{Results}\label{sec:results}
\subsection{Distinct Strengths of System 1 \& System 2 Models}\label{res:system2}

\Cref{tab:compare} shows a comparison of exact matching accuracy across 13 benchmarks for Llama and Mistral. Specifically, we compare the base models with the $\mathcal{S}1$ and $\mathcal{S}2$ variants, and include results for CoT prompting for reference. Our findings reveal distinct performance trends for the $\mathcal{S}1$ and $\mathcal{S}2$ models, highlighting their respective strengths in different reasoning benchmarks.

In all arithmetic benchmarks (MultiArith, GSM8K, AddSub, AQuA, SingleEq, and SVAMP), $\mathcal{S}2$ models consistently outperformed both the base models and their $\mathcal{S}1$ counterparts. This improvement is most significant in the AddSub and SingleEq benchmarks. Similarly, $\mathcal{S}2$ models outperformed $\mathcal{S}1$ models in nearly all symbolic reasoning benchmarks (Coin and Letter), which require pattern recognition and logical structuring, further validating the idea that deliberative, slow-thinking models enhance performance in structured reasoning. 
While both approaches achieve high accuracy, $\mathcal{S}1$’s reliance on heuristic shortcuts introduce small but systematic errors that $\mathcal{S}2$’s deliberate, stepwise computations tend to avoid, such as rounding the number or adding numbers without checking. These findings are further supported by our AddSub analysis in \Cref{appendix:failure_patterns}.

\begin{table*}[t]
\centering
\footnotesize
\caption{Accuracy comparison of our System 1 and System 2-aligned models against instruction-tuned and CoT baselines across benchmarks. Each cell shows accuracy, with parentheses indicating the difference from the baseline. Color intensity reflects the magnitude of deviation.}
\label{tab:compare}
\resizebox{\textwidth}{!}{ 
\begin{tabular}{ll|*{6}{p{1.5cm}<{\centering}}|*{2}{p{1.5cm}<{\centering}}|*{5}{p{1.5cm}<{\centering}}}
\multicolumn{2}{c}{} & \multicolumn{6}{c}{\textbf{Arithmetic}} & \multicolumn{2}{c}{\textbf{Symbolic}} &\multicolumn{5}{c}{\textbf{Common Sense}} \\
& & \textbf{MultiArith} & \textbf{GSM8K} & \textbf{AddSub} & \textbf{AQuA} & \textbf{SingleEq} & \textbf{SVAMP} & \textbf{Coin} & \textbf{Letter} & \textbf{CSQA} & \textbf{Strategy} & \textbf{PIQA} & \textbf{SIQA} & \textbf{COM2SENSE} \\
\midrule
\multirow[c]{2}{*}{\rotatebox{90}{\textbf{System 2}}} & \textbf{DPO} & 
\highlightcell{\makecell{98.67\\(+1.0)}}{+1.0} & 
\highlightcell{\makecell{79.37\\(+0.88)}}{+0.88} & 
\highlightcell{\makecell{89.87\\(+7.4)}}{+7.4} & 
\highlightcell{\makecell{49.21\\(+0.39)}}{+0.39} & 
\highlightcell{\makecell{94.37\\(+3.65)}}{+3.65} & 
\highlightcell{\makecell{85.4\\(+4.9)}}{+4.9} & 
\highlightcell{\makecell{93.8 \\ (-0.4)}}{-0.4} & 
\highlightcell{\makecell{86.2 \\ (+2.2)}}{+2.2} & 
\highlightcell{\makecell{71.42 \\ (0)}}{0} & 
\highlightcell{\makecell{60.87 \\ (-6.68)}}{-6.68} &
\highlightcell{\makecell{81.15 \\ (-2.01)}}{-1.38} &
\highlightcell{\makecell{67.93 \\ (-3.19)}}{-3.19} &
\highlightcell{\makecell{76.42 \\ (-2.6)}}{-2.6}
 \\
& \textbf{SIMPO} & 
\highlightcell{\makecell{97.83\\(+0.16)}}{+0.16} & 
\highlightcell{\makecell{79.38\\(+0.89)}}{+0.89} & 
\highlightcell{\makecell{90.13\\(+7.66)}}{+7.66} & 
\highlightcell{\makecell{54.72\\(+6.78)}}{+6.78} & 
\highlightcell{\makecell{94.49\\(+3.77)}}{+3.77} & 
\highlightcell{\makecell{81.7\\(+1.2)}}{+1.2} & 
\highlightcell{\makecell{94.4\\(+0.2)}}{+0.2} & 
\highlightcell{\makecell{84.8\\(+0.8)}}{+0.8} & 
\highlightcell{\makecell{69.62 \\(-1.8)}}{-1.8} & 
\highlightcell{\makecell{67.38 \\ (-0.17)}}{-0.17} &
\highlightcell{\makecell{81.49 \\ (-1.67)}}{-1.67} &
\highlightcell{\makecell{69.16 \\ (-1.96)}}{-1.96} &
\highlightcell{\makecell{78.21 \\ (-0.81)}}{-0.81}
 \\
 
\multicolumn{2}{c}{\textbf{Llama-3}}  & 
97.67 & 78.49 & 82.47 & 48.82 & 90.72 & 80.5 & 94.2 & 84 & 71.42 & 67.55 & 83.16 & 71.12 & 79.02\\

\multicolumn{2}{c}{\textbf{Llama-3-CoT}}  & 
97.83 & 78.54 & 82.03 & 49.21 & 88.19 & 80.9 & 94.8 & 84.2 & 71.58 & 67.38& 83.34 & 70.97& 79.86\\

\multirow[c]{2}{*}{\rotatebox{90}{\textbf{System 1}}} & \textbf{DPO} & 
\highlightcell{\makecell{98.5\\(+0.83)}}{+0.83} & 
\highlightcell{\makecell{77.01\\(-1.48)}}{-1.48} & 
\highlightcell{\makecell{80.76\\(-1.71)}}{-1.71} & 
\highlightcell{\makecell{46.46\\(-2.36)}}{-2.36} & 
\highlightcell{\makecell{77.24\\(-13.48)}}{-13.48} & 
\highlightcell{\makecell{78\\(-2.5)}}{-2.5} & 
\highlightcell{\makecell{93.4\\(-0.8)}}{-0.8} & 
\highlightcell{\makecell{83.8\\(-0.2)}}{-0.2} & 
\highlightcell{\makecell{72.81 \\ (+1.39)}}{+1.39} & 
\highlightcell{\makecell{68.21 \\ (+0.66)}}{+0.66} &
\highlightcell{\makecell{83.94 \\ (+0.78)}}{+0.78} &
\highlightcell{\makecell{72.16 \\ (+1.04)}}{+1.04} &
\highlightcell{\makecell{79.99 \\ (+0.97)}}{+0.97}
\\
& \textbf{SIMPO} & 
\highlightcell{\makecell{97.5\\(-0.17)}}{-0.17} & 
\highlightcell{\makecell{77.79\\(-0.7)}}{-0.7} & 
\highlightcell{\makecell{80.51\\(-1.96)}}{-1.96} & 
\highlightcell{\makecell{48.03\\(-0.79)}}{-0.79} & 
\highlightcell{\makecell{87.4\\(-3.32)}}{-3.32} & 
\highlightcell{\makecell{79.3\\(-1.2)}}{-1.2} & 
\highlightcell{\makecell{90\\(-4.2)}}{-4.2} & 
\highlightcell{\makecell{83.8\\(-0.2)}}{-0.2} & 
\highlightcell{\makecell{72.32 \\(+0.9)}}{+0.9} & 
\highlightcell{\makecell{67.73 \\ (+0.18)}}{+0.18} &
\highlightcell{\makecell{83.35 \\ (+0.19)}}{+0.19} &
\highlightcell{\makecell{71.67 \\ (+0.55)}}{+0.55} &
\highlightcell{\makecell{81.46 \\ (+2.44)}}{+2.44}
 \\

\midrule

\multirow[c]{2}{*}{\rotatebox{90}{ \textbf{System 2}}} & \textbf{DPO} & 
\highlightcell{\makecell{78.83\\(+1.16)}}{+1.16} & 
\highlightcell{\makecell{56.45\\(+1.47)}}{+1.47} & 
\highlightcell{\makecell{81.27\\(+6.79)}}{+6.79} & 
\highlightcell{\makecell{32.68\\(+1.19)}}{+1.19} & 
\highlightcell{\makecell{84.84 \\(+0.98)}}{+0.98} & 
\highlightcell{\makecell{69.1\\(+3.4)}}{+3.4} & 
\highlightcell{\makecell{41 \\(-2.2)}}{-2.2} & 
\highlightcell{\makecell{8.6\\(+8)}}{+8} & 
\highlightcell{\makecell{62.82 \\ (-3.44)}}{-3.44} & 
\highlightcell{\makecell{56.81 \\ (-8.6)}}{-8.6} &
\highlightcell{\makecell{80.49\\(0) }}{0} &
\highlightcell{\makecell{57.77\\ (-2.24)}}{-2.24} &
\highlightcell{\makecell{66.73 \\ (-1.64)}}{-1.64}
 \\
& \textbf{SIMPO} & 
\highlightcell{\makecell{78.3\\(+0.63)}}{+0.63} & 
\highlightcell{\makecell{55.42\\(+0.53)}}{+0.53} & 
\highlightcell{\makecell{82.28\\(+7.8)}}{+7.8} & 
\highlightcell{\makecell{34.25\\(+2.76)}}{+2.76} & 
\highlightcell{\makecell{86.81\\(+2.95)}}{+2.95} & 
\highlightcell{\makecell{68.5\\(+2.8)}}{+2.8} & 
\highlightcell{\makecell{45.4\\(+2.2)}}{+2.2} & 
\highlightcell{\makecell{7.8\\(+6.2)}}{+6.2} & 
\highlightcell{\makecell{64.78 \\(-1.48)}}{-1.48} & 
\highlightcell{\makecell{63.75 \\ (-1.66)}}{-1.66} &
\highlightcell{\makecell{82.07\\ (-0.46)}}{-0.46} &
\highlightcell{\makecell{59.82\\ (-0.19)}}{-0.19} &
\highlightcell{\makecell{68.15 \\ (-0.22)}}{-0.22}
 \\
\multicolumn{2}{c}{\textbf{Mistral}}  & 
77.67 & 54.89 & 79.75 & 31.49 & 83.86 & 66.26 & 43.2 & 1.6 & 66.26 & 65.41 & 82.53 & 60.01 &68.37\\
\multicolumn{2}{c}{\textbf{Mistral-CoT}}  & 
 78.3 & 54.96 & 80.25 & 33.07 & 83.66 & 67.8 & 43.8 & 1.6 & 66.18 & 65.49 & 82.21& 60.76& 69.01\\

\multirow[c]{3}{*}{\rotatebox{90}{ \textbf{System 1}}} & \textbf{DPO} & 
\highlightcell{\makecell{77.5\\(-0.17)}}{-0.17} & 
\highlightcell{\makecell{51.4\\(-3.49)}}{-3.49} & 
\highlightcell{\makecell{79.49\\(-0.26)}}{-0.26} & 
\highlightcell{\makecell{29.53\\(-1.96)}}{-1.96} & 
\highlightcell{\makecell{83.07\\(-0.79)}}{-0.79} & 
\highlightcell{\makecell{67.4\\(-0.2)}}{-0.2} & 
\highlightcell{\makecell{40.4\\(-2.8)}}{-2.8} & 
\highlightcell{\makecell{0\\(-1.6)}}{-1.6} & 
\highlightcell{\makecell{67.4 \\ (+1.14)}}{+1.14} & 
\highlightcell{\makecell{65.49 \\ (+0.08)}}{+0.08} &
\highlightcell{\makecell{83.22 \\ (+0.69) }}{+0.69} &
\highlightcell{\makecell{60.01 \\ (0)}}{0} &
\highlightcell{\makecell{70.83\\ (+2.46)}}{+2.46}
\\
& \textbf{SIMPO} & 
\highlightcell{\makecell{77\\(-0.67)}}{-0.67} & 
\highlightcell{\makecell{53.61\\(-1.28)}}{-1.28} & 
\highlightcell{\makecell{78.73\\(-1.02)}}{-1.02} & 
\highlightcell{\makecell{31.1\\(-0.39)}}{-0.39} & 
\highlightcell{\makecell{83.67\\(-0.19)}}{-0.19} & 
\highlightcell{\makecell{67.3\\(-0.3)}}{-0.3} & 
\highlightcell{\makecell{43\\(-0.2)}}{-0.2} & 
\highlightcell{\makecell{0\\(-1.6)}}{-1.6} & 
\highlightcell{\makecell{67.32 \\(+1.06)}}{+1.06} & 
\highlightcell{\makecell{65.51 \\ (+0.1)}}{+0.1} &
\highlightcell{\makecell{82.84 \\(+1.31)}}{+1.31} &
\highlightcell{\makecell{60.93 \\ (+0.92)}}{+0.92} &
\highlightcell{\makecell{69.13 \\ (+0.76)}}{+0.76}
 \\
\bottomrule
\end{tabular}
}

\end{table*}

Conversely, $\mathcal{S}1$ models consistently excelled all of their $\mathcal{S}2$ counterparts, the base models, and the CoT variant on all commonsense reasoning benchmarks (CSQA, StrategyQA, PIQA, SIQA, and COM2SENSE), which depend on intuitive judgments and heuristic shortcuts. 
While $\mathcal{S}2$ reasoning is correct, its more deliberate nature can often lead to overthinking, producing overly cautious or extensively interpretive responses that diverge from typical human reactions in rapid, intuitive situations. For example, when asked what a kindergarten teacher does before nap time, $\mathcal{S}2$ suggests ``encourage quiet behavior'' instead of ``tell a story,'' or predicts ``laughter'' rather than ``fight'' if you surprise an angry person. As shown in Appendix \ref{appendix:failure_patterns}, this tendency to favor completeness over contextual fit makes $\mathcal{S}2$ less reliable for quick, socially grounded tasks.

Llama models generally outperformed Mistral across all benchmarks, suggesting stronger foundational reasoning capabilities further enhanced by $\mathcal{S}1$ and $\mathcal{S}2$ alignment. Moreover, instruction-tuned models with CoT prompts exhibited marginal gains over their base counterparts, as step-by-step reasoning is already internalized during pretraining on CoT data \citep{llama3modelcard}. Accordingly, we adopt the base Llama model as our primary baseline in subsequent experiments.

In summary, our results showcase that $\mathcal{S}2$ models excel in structured, multi-step reasoning such as arithmetic and symbolic reasoning, while $\mathcal{S}1$ models are effective in intuitive and commonsense reasoning benchmarks. These findings highlight the significant potential of dual-process alignment for boosting LLM performance across a diverse range of reasoning paradigms.

\subsection{Length Differences Across Reasoning Styles}\label{res:system1}

A recent trend in LLM performance, exemplified by models such as DeepSeek R1 \citep{guo2025deepseek}, is that achieving stronger benchmark results often correlates with producing longer reasoning chains, even if not explicitly trained to do so. This correlation raises the question of whether such verbose responses truly reflect enhanced reasoning capabilities or if they are simply a formatting artifact of current high-performing models. In our studies, this concern is particularly relevant for $\mathcal{S}2$ models, which are expected to behave more deliberatively. To investigate this, we analyze output lengths with the two-stage prompting setup described in \Cref{experiment_setup}.

\begin{wrapfigure}{r}{0.54\textwidth} 
\includegraphics[width=\textwidth]{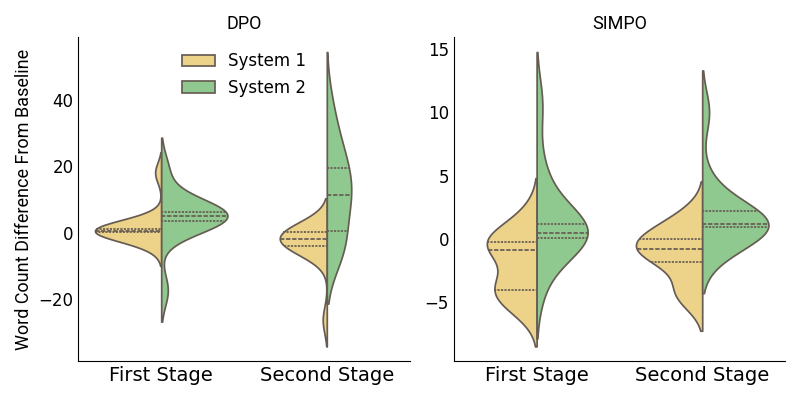} 
\caption{Token difference between System 1 and System 2 responses relative to Llama model across prompting stages and alignment methods.}
\label{fig:resanaylsis}
\end{wrapfigure}

As shown in \Cref{fig:resanaylsis}, $\mathcal{S}2$-aligned models generate significantly longer responses than their $\mathcal{S}1$ counterparts, relative to the Llama baseline, under both alignment methods, DPO ($t(8836) = 57.14$, $p < .001$) and SimPO ($t(8586) = 9.833$, $p < .001$). This difference emerges specifically in the second stage, where models are prompted to finalize their responses, while response lengths remain comparable in the first stage, where both models are simply instructed to reason. Although both models were trained on equal-length preference pairs as described in \Cref{sec:dataset}, $\mathcal{S}2$ models still tend to elaborate more during finalization, consistent with their alignment toward deliberative reasoning.

While longer reasoning chains are often associated with stronger performance, our findings suggest that this extended reasoning can also introduce inefficiencies or even degrade quality in contexts where concise, heuristic-driven reasoning is more appropriate. In particular, tasks requiring commonsense or intuitive judgments are often better handled by $\mathcal{S}1$ models, which respond more directly. 
\pr{This aligns with emerging work on ``overthinking'' phenomenon, where excessive deliberation hurts performance \citep{chen2024not, cuadron2025danger}.}
Overall, extended reasoning is not universally beneficial, and reasoning strategies must be evaluated in relation to the task.

\begin{figure*}[htbp]
\centering
  \includegraphics[width=\columnwidth]{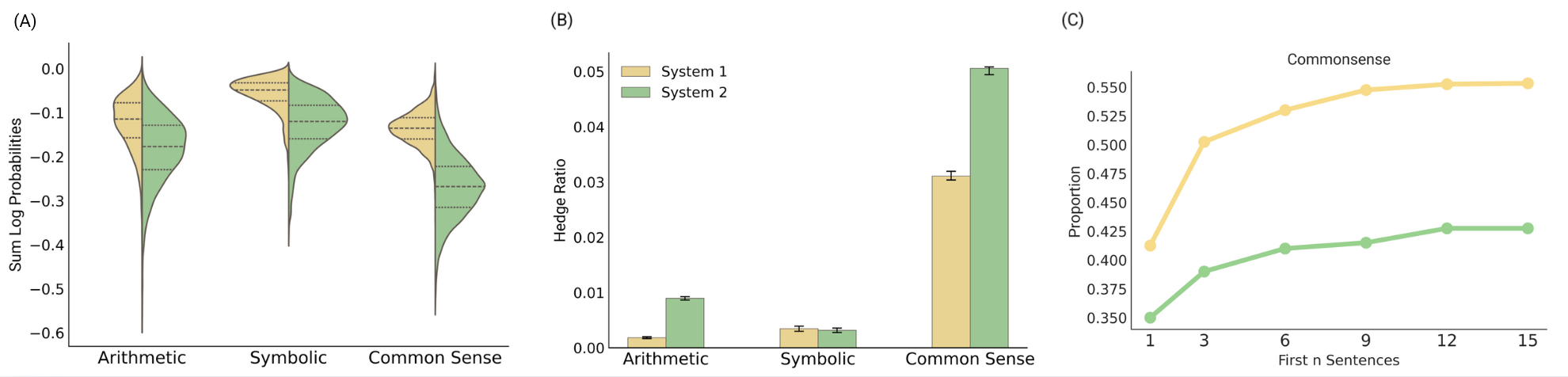}
  \caption{(A) Log probabilities of models' reasoning indicating internal uncertainty; 
  (B) Hedge word ratio showing surface-level uncertainty; 
  (C) Proportion of definitive answers in the first n sentences.}
  \label{fig:exp4}
\end{figure*}

\subsection{Uncertainty across Reasoning Styles}\label{res:uncertainty}
 A key insight from psychology and neuroscience is that $\mathcal{S}1$ operates on confident heuristics, providing quick, intuitive judgments, while $\mathcal{S}2$ engages in more deliberate, analytical thought, accurately assessing the uncertainty associated with its conclusions \citep{daw2005uncertainty, lee2014neural, keramati2011speed, xu2021career}. To examine uncertainty and confidence, we consider three different characteristics: 1) token-level uncertainty; 2) the presence of hedge words in model output \citep{lakoff1973hedges, ott2018hedging}; and 3) definitive commitment to responses in $\mathcal{S}1$ versus $\mathcal{S}2$.

Plot A in \Cref{fig:exp4} shows that $\mathcal{S}2$ models consistently generate tokens with lower confidence than $\mathcal{S}1$ models, based on token-level uncertainty from logits.
This trend holds across arithmetic $t(4075) = 54.53, p < .001$, symbolic $t(999) = 42.53, p < .001$, and commonsense $t(3510) = 106.86, p < .001$ benchmarks. Additionally, we analyzed surface-level uncertainty in model reasoning by examining word choices. \Cref{fig:exp4}, Plot B shows $\mathcal{S}2$-aligned models use significantly more hedge words, in arithmetic $t(4075) = 22.03, p < .001$ and commonsense $t(3510) = 21.49, p < .001$ when models reiterate their reasoning. While increased uncertainty enhances analytical reasoning, it may hinder tasks requiring rapid, intuitive judgments. To assess early-stage response conclusiveness, we used LLM-as-Judge \citep{zheng2023judging} as detailed in \Cref{appendix:additional-insights-into-models-reasoning}. \Cref{fig:exp4}, Plot C shows $\mathcal{S}1$ models provide significantly more definitive responses than $\mathcal{S}2$ models in commonsense reasoning, $\textit{McNemar's } \chi^2(1, 400) = 20.0, p < .001$, regardless of where in the response the definitive responses are reached (see \Cref{appendix:additional-insights-into-models-reasoning}).

This analysis reinforces the idea that different reasoning styles are suited to different tasks. Greater uncertainty in models' generated reasoning suggests that $\mathcal{S}2$ models can explore alternative reasoning paths more effectively. This uncertainty is reflected in both their model output probabilities and word choices. $\mathcal{S}2$ models' superior performance in arithmetic benchmarks highlights the benefits of deliberate, effortful processing in tasks that demand exploration and uncertainty. 
On the other hand, the greater tendency of $\mathcal{S}1$ models to commit to responses in a more definitive way aligns with their advantage in tasks requiring rapid and intuitive judgments. This behavior is observed exclusively in commonsense reasoning, where quick, decisive responses are advantageous—a trend supported by human studies \citep{byrd2022bounded} and confirmed by our findings in \Cref{res:system2}. However, it does not appear in other benchmarks (see \Cref{appendix:additional-insights-into-models-reasoning}), suggesting that the activation of a particular reasoning style is context-dependent and influenced by task demands. 

\subsection{Moving from Fast to Slow Thinking}\label{res:ratio}
\begin{figure*}
\centering
  \includegraphics[width=0.95\columnwidth]{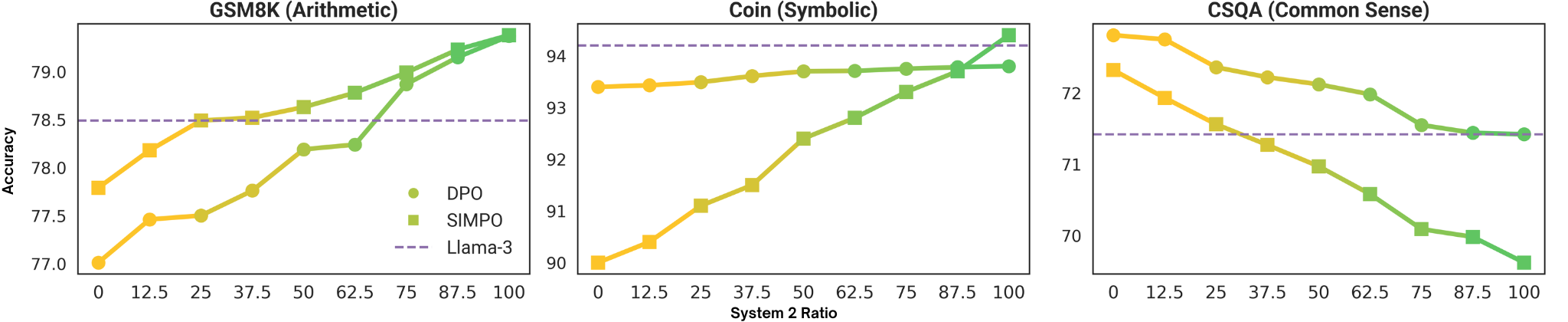}
  \caption{Accuracy across benchmark categories as reasoning shifts from System 1 to System 2.}
  \label{fig:ratio}
\end{figure*}

In the previous analysis, $\mathcal{S}1$ and $\mathcal{S}2$ models can be viewed as endpoints of a broader spectrum of reasoning strategies. Paralleling approaches in cognitive psychology \citep{daw2011model, piray2021linear}, we explored this spectrum by constructing interpolated models—blending $\mathcal{S}1$ and $\mathcal{S}2$ preferred answers at varying ratios in the alignment dataset.
\Cref{fig:ratio} demonstrates a consistent, monotonic transition in accuracy across representative benchmarks from three reasoning categories (all $r^2 > 0.9, p < 0.001$), a pattern visible across all benchmarks (see \Cref{appendix:full-ratio}). While arithmetic and symbolic reasoning benchmarks exhibit a steady increase in accuracy moving toward $\mathcal{S}2$ thinking, commonsense reasoning benchmarks show the opposite trend, with accuracy increasing as models rely more on $\mathcal{S}1$ reasoning. This trade-off highlights that both reasoning styles offer unique advantages, with $\mathcal{S}2$ excelling in structured, multi-step problem-solving and $\mathcal{S}1$ providing efficient, adaptable responses in intuitive scenarios. These findings strengthen the importance of task-dependent reasoning strategies that leverage the strengths of both $\mathcal{S}1$ and $\mathcal{S}2$ thinking. Critically, there are no sudden drops or fluctuations in performance when transitioning between reasoning styles. This stability indicates that the shift from $\mathcal{S}1$ to $\mathcal{S}2$ reasoning is gradual and predictable, without any unexpected anomalies. This observation reinforces the idea that LLMs can be strategically guided toward different reasoning styles, allowing for more adaptive problem-solving.

\subsection{Entropy-Guided Model Selection}

\pr{
We evaluated the dynamic model proposed in \Cref{sec:dynamicmethod} on our 13 reasoning benchmarks, varying the weight $w$ in \Cref{eq:combined_score}. Overall, the dynamic models consistently outperform their base counterparts across the different alignment algorithms on nearly all benchmarks. The best performance was achieved with $w=0.4$, under which the Llama DPO-dynamic model achieved higher accuracy than the base model on 12 of the 13 benchmarks, while the SimPO-dynamic version improved on 11 benchmarks. Given the significance of this finding, we also replicated the analysis with Mistral models, where the DPO-dynamic model outperformed the base on 11 of 13 benchmarks, while the SimPO-dynamic model improved on 12 of 13 benchmarks (see \Cref{appendix:dynamicappendixres}).
}

\begin{figure*}[htbp]

    \centering

    \includegraphics[width=\linewidth]{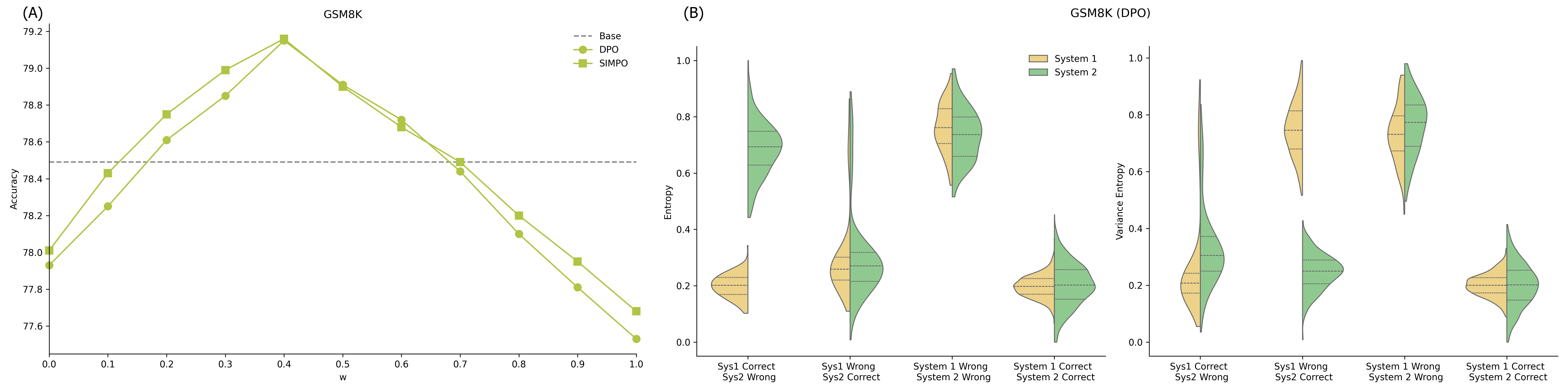}

\caption{(A) Performance of Llama models (DPO- and SimPO-dynamic models) on the GSM8K dataset as $w$ varies in \Cref{eq:combined_score}. The dashed line represents the accuracy of the base Llama model. (B) Violin plots of average entropy ($\bar{H}$) and its variance ($\sigma^2$) distribution for DPO-aligned Llama models on GSM8K, broken down by four possible outcomes.}
    \label{fig:one-dynamic-plot}
\end{figure*}

\pr{
Furthermore, to validate the balance between uncertainty ($\hat{H}$) and instability ($\hat{\sigma}^2$) in our dynamic model, we analyzed the distributions of $\bar{H}$  and $\sigma^2$  between the two systems. As an illustration, \Cref{fig:one-dynamic-plot} shows GSM8k accuracy across different $w$ values alongside the corresponding entropy statistics; results for the remaining benchmarks are provided in \Cref{appendix:dynamicappendixres} and follow the same trend. This analysis reveals systematic differences between correct and incorrect responses in $\mathcal{S}1$ and $\mathcal{S}2$ models. In general, high $\bar{H}$ in either system is associated with incorrect responses, whereas for both correct and incorrect cases the two systems exhibit very similar entropy statistics. We also observe that $\hat{H}$ is generally lower for $\mathcal{S}1$ models, indicating greater confidence, while $\hat{\sigma}^2$ is lower for $\mathcal{S}2$ models, indicating greater stability. These findings are consistent with \Cref{res:uncertainty} and with prior research in psychology and neuroscience. Together, this analysis provides empirical justification for using entropy signals as the basis of our scoring method in \Cref{sec:dynamicmethod}.
}

\vspace{-4mm}

\section{Conclusion}\label{sec:conclusion}
A central question in current LLM development is whether structured, step-by-step reasoning is always beneficial, or whether a more flexible range of reasoning strategies is needed. Inspired by dual-process theories of human cognition, we studied LLMs explicitly aligned with $\mathcal{S}1$ and $\mathcal{S}2$ thinking, representing fast, confident, heuristic reasoning and slow, analytical reasoning, respectively. Our findings indicate that, much like in human cognition, reasoning in LLMs is not a one-size-fits-all solution: different reasoning modes are effective in different contexts and downstream tasks. $\mathcal{S}2$ excels in arithmetic and symbolic reasoning, while $\mathcal{S}1$ is more effective and accurate in commonsense reasoning (\Cref{res:system2}). Additionally, $\mathcal{S}1$ models generate responses with fewer tokens, highlighting its efficiency in decision-making (\Cref{res:system1}). Our analysis in \Cref{res:uncertainty} illustrated that $\mathcal{S}2$ models exhibit greater uncertainty throughout the reasoning process, potentially resulting them to engage in more structured, step-by-step problem-solving. In contrast, $\mathcal{S}1$ models display higher confidence, allowing them to reach responses faster, which is particularly advantageous for tasks requiring rapid, intuitive judgments. \pr{Moreover}, training intermediate models with blended ratios of preferred $\mathcal{S}1$ and $\mathcal{S}2$ responses revealed smooth, monotonic shifts in performance across benchmarks (\Cref{res:ratio}), supporting the view that LLM reasoning should lie on a continuous, tunable spectrum rather than a binary divide. \pr{Finally, we proposed a dynamic model that selects adaptively between $\mathcal{S}1$ and $\mathcal{S}2$ reasoning based on entropy signals. Remarkably, this method requires no additional training yet consistently improves performance across diverse reasoning benchmarks. This demonstrates that simple uncertainty-based arbitration can output the most reliable response.}

Beyond these empirical findings, our study aligns with broader principles observed across cognitive science and neuroscience. The observation that $\mathcal{S}1$ models generate faster and more confident responses echoes established theories in human cognition, where intuitive, heuristic thinking allows for rapid decision-making. Similarly, the higher uncertainty exhibited by $\mathcal{S}2$ models aligns with neuroscience findings that deliberate reasoning involves greater cognitive load, self-monitoring, and exploring more paths. These parallels suggest that LLMs, when properly aligned, can mirror key aspects of human cognition, offering new insights into both artificial and natural intelligence.

Our work bridges between LLM development and cognitive science, highlighting efficiency-accuracy trade-offs in LLMs, similar to those long observed in human cognition. We align models with reasoning behaviors that follow well-known cognitive heuristics, which humans use in everyday thinking, like $\mathcal{S}1$'s rapid, intuitive judgments and $\mathcal{S}2$'s deliberate, analytical thought, and show they can follow the dynamic interplay between fast and slow thinking. This alignment not only informs more sophisticated training and evaluation strategies but also suggests that future LLMs can be designed to possess a more cognitively grounded flexibility, allowing them to adapt their reasoning as effectively as humans do when faced with diverse task demands. Finally, models that reason in ways that are cognitively interpretable, mirroring the human brain’s strategies for learning, decision making, and inference, may also be more predictable, steerable, and trustworthy in deployment. In this light, dual-process alignment connects cognitive science and neuroscience with model capabilities, enabling future LLMs to reason more like humans, not just in what they conclude, but in how they get there.

This paper takes a first step toward adaptive reasoning in LLMs, enabling dynamic shift between heuristic and deliberative thinking based on task demands. Furthermore, understanding how to optimally balance speed and accuracy in LLMs can have significant implications for real-world applications, from conversational agents to automated decision-making systems. In practice, this approach allows deliberate trade-offs between answer quality and response speed, using fewer reasoning steps when time is critical.

\bibliographystyle{iclr2026_conference}
\bibliography{iclr2026_conference}

\begin{thebibliography}{96}
\providecommand{\natexlab}[1]{#1}
\providecommand{\url}[1]{\texttt{#1}}
\expandafter\ifx\csname urlstyle\endcsname\relax
  \providecommand{\doi}[1]{doi: #1}\else
  \providecommand{\doi}{doi: \begingroup \urlstyle{rm}\Url}\fi

\bibitem[Abdin et~al.(2024)Abdin, Aneja, Behl, Bubeck, Eldan, Gunasekar, Harrison, Hewett, Javaheripi, Kauffmann, et~al.]{abdin2024phi}
Marah Abdin, Jyoti Aneja, Harkirat Behl, S{\'e}bastien Bubeck, Ronen Eldan, Suriya Gunasekar, Michael Harrison, Russell~J Hewett, Mojan Javaheripi, Piero Kauffmann, et~al.
\newblock Phi-4 technical report.
\newblock \emph{arXiv preprint arXiv:2412.08905}, 2024.

\bibitem[AI@Meta(2024)]{llama3modelcard}
AI@Meta.
\newblock Llama 3 model card.
\newblock \emph{arXiv preprint}, 2024.
\newblock URL \url{https://github.com/meta-llama/llama3/blob/main/MODEL_CARD.md}.

\bibitem[Alizadeh et~al.(2023)Alizadeh, Kubli, Samei, Dehghani, Bermeo, Korobeynikova, and Gilardi]{alizadeh2023open}
Meysam Alizadeh, Ma{\"e}l Kubli, Zeynab Samei, Shirin Dehghani, Juan~Diego Bermeo, Maria Korobeynikova, and Fabrizio Gilardi.
\newblock Open-source large language models outperform crowd workers and approach chatgpt in text-annotation tasks.
\newblock \emph{arXiv preprint arXiv:2307.02179}, 42, 2023.

\bibitem[Balleine \& Dickinson(1998)Balleine and Dickinson]{balleine1998goal}
Bernard~W Balleine and Anthony Dickinson.
\newblock Goal-directed instrumental action: contingency and incentive learning and their cortical substrates.
\newblock \emph{Neuropharmacology}, 37\penalty0 (4-5):\penalty0 407--419, 1998.

\bibitem[Bisk et~al.(2020)Bisk, Zellers, Bras, Gao, and Choi]{bisk2020piqa}
Yonatan Bisk, Rowan Zellers, Ronan~Le Bras, Jianfeng Gao, and Yejin Choi.
\newblock Piqa: Reasoning about physical commonsense in natural language.
\newblock In \emph{Thirty-Fourth AAAI Conference on Artificial Intelligence}, 2020.

\bibitem[Booch et~al.(2021)Booch, Fabiano, Horesh, Kate, Lenchner, Linck, Loreggia, Murgesan, Mattei, Rossi, et~al.]{booch2021thinking}
Grady Booch, Francesco Fabiano, Lior Horesh, Kiran Kate, Jonathan Lenchner, Nick Linck, Andreas Loreggia, Keerthiram Murgesan, Nicholas Mattei, Francesca Rossi, et~al.
\newblock Thinking fast and slow in ai.
\newblock In \emph{Proceedings of the AAAI Conference on Artificial Intelligence}, volume~35, pp.\  15042--15046, 2021.

\bibitem[Brown et~al.(2020)Brown, Mann, Ryder, Subbiah, Kaplan, Dhariwal, Neelakantan, Shyam, Sastry, Askell, et~al.]{brown2020language}
Tom Brown, Benjamin Mann, Nick Ryder, Melanie Subbiah, Jared~D Kaplan, Prafulla Dhariwal, Arvind Neelakantan, Pranav Shyam, Girish Sastry, Amanda Askell, et~al.
\newblock Language models are few-shot learners.
\newblock \emph{Advances in neural information processing systems}, 33:\penalty0 1877--1901, 2020.

\bibitem[Byrd(2022)]{byrd2022bounded}
Nick Byrd.
\newblock Bounded reflectivism and epistemic identity.
\newblock \emph{Metaphilosophy}, 53\penalty0 (1):\penalty0 53--69, 2022.

\bibitem[Chen et~al.(2024)Chen, Xu, Liang, He, Pang, Yu, Song, Liu, Zhou, Zhang, et~al.]{chen2024not}
Xingyu Chen, Jiahao Xu, Tian Liang, Zhiwei He, Jianhui Pang, Dian Yu, Linfeng Song, Qiuzhi Liu, Mengfei Zhou, Zhuosheng Zhang, et~al.
\newblock Do not think that much for 2+ 3=? on the overthinking of o1-like llms.
\newblock \emph{arXiv preprint arXiv:2412.21187}, 2024.

\bibitem[Chung et~al.(2024)Chung, Hou, Longpre, Zoph, Tay, Fedus, Li, Wang, Dehghani, Brahma, et~al.]{chung2024scaling}
Hyung~Won Chung, Le~Hou, Shayne Longpre, Barret Zoph, Yi~Tay, William Fedus, Yunxuan Li, Xuezhi Wang, Mostafa Dehghani, Siddhartha Brahma, et~al.
\newblock Scaling instruction-finetuned language models.
\newblock \emph{Journal of Machine Learning Research}, 25\penalty0 (70):\penalty0 1--53, 2024.

\bibitem[Cobbe et~al.(2021)Cobbe, Kosaraju, Bavarian, Chen, Jun, Kaiser, Plappert, Tworek, Hilton, Nakano, Hesse, and Schulman]{cobbe2021trainingverifierssolvemath}
Karl Cobbe, Vineet Kosaraju, Mohammad Bavarian, Mark Chen, Heewoo Jun, Lukasz Kaiser, Matthias Plappert, Jerry Tworek, Jacob Hilton, Reiichiro Nakano, Christopher Hesse, and John Schulman.
\newblock Training verifiers to solve math word problems, 2021.
\newblock URL \url{https://arxiv.org/abs/2110.14168}.

\bibitem[Cuadron et~al.(2025)Cuadron, Li, Ma, Wang, Wang, Zhuang, Liu, Schroeder, Xia, Mao, et~al.]{cuadron2025danger}
Alejandro Cuadron, Dacheng Li, Wenjie Ma, Xingyao Wang, Yichuan Wang, Siyuan Zhuang, Shu Liu, Luis~Gaspar Schroeder, Tian Xia, Huanzhi Mao, et~al.
\newblock The danger of overthinking: Examining the reasoning-action dilemma in agentic tasks.
\newblock \emph{arXiv preprint arXiv:2502.08235}, 2025.

\bibitem[Daw et~al.(2005)Daw, Niv, and Dayan]{daw2005uncertainty}
Nathaniel~D Daw, Yael Niv, and Peter Dayan.
\newblock Uncertainty-based competition between prefrontal and dorsolateral striatal systems for behavioral control.
\newblock \emph{Nature neuroscience}, 8\penalty0 (12):\penalty0 1704--1711, 2005.

\bibitem[Daw et~al.(2011)Daw, Gershman, Seymour, Dayan, and Dolan]{daw2011model}
Nathaniel~D Daw, Samuel~J Gershman, Ben Seymour, Peter Dayan, and Raymond~J Dolan.
\newblock Model-based influences on humans' choices and striatal prediction errors.
\newblock \emph{Neuron}, 69\penalty0 (6):\penalty0 1204--1215, 2011.

\bibitem[Del{\'{e}}tang et~al.(2023)Del{\'{e}}tang, Ruoss, Grau{-}Moya, Genewein, Wenliang, Catt, Cundy, Hutter, Legg, Veness, and Ortega]{deletang2023neural}
Gr{\'{e}}goire Del{\'{e}}tang, Anian Ruoss, Jordi Grau{-}Moya, Tim Genewein, Li~Kevin Wenliang, Elliot Catt, Chris Cundy, Marcus Hutter, Shane Legg, Joel Veness, and Pedro~A. Ortega.
\newblock Neural networks and the chomsky hierarchy.
\newblock In \emph{11th International Conference on Learning Representations}, 2023.

\bibitem[Deng et~al.(2024)Deng, Qiu, Tan, Qu, Pan, Cheng, Xu, and Chu]{deng2024cognidual}
Yongxin Deng, Xihe Qiu, Xiaoyu Tan, Chao Qu, Jing Pan, Yuan Cheng, Yinghui Xu, and Wei Chu.
\newblock Cognidual framework: Self-training large language models within a dual-system theoretical framework for improving cognitive tasks.
\newblock \emph{arXiv preprint arXiv:2409.03381}, 2024.

\bibitem[Dolan \& Dayan(2013)Dolan and Dayan]{dolan2013goals}
Ray~J Dolan and Peter Dayan.
\newblock Goals and habits in the brain.
\newblock \emph{Neuron}, 80\penalty0 (2):\penalty0 312--325, 2013.

\bibitem[Dumpala et~al.(2024)Dumpala, Jaiswal, Shama~Sastry, Milios, Oore, and Sajjad]{dumpala2024sugarcrepe++}
Sri~Harsha Dumpala, Aman Jaiswal, Chandramouli Shama~Sastry, Evangelos Milios, Sageev Oore, and Hassan Sajjad.
\newblock Sugarcrepe++ dataset: Vision-language model sensitivity to semantic and lexical alterations.
\newblock \emph{Advances in Neural Information Processing Systems}, 37:\penalty0 17972--18018, 2024.

\bibitem[Echterhoff et~al.(2024)Echterhoff, Liu, Alessa, McAuley, and He]{echterhoff2024cognitive}
Jessica Echterhoff, Yao Liu, Abeer Alessa, Julian McAuley, and Zexue He.
\newblock Cognitive bias in decision-making with llms.
\newblock In \emph{Findings of the Association for Computational Linguistics: EMNLP 2024}, pp.\  12640--12653, 2024.

\bibitem[Evans \& Stanovich(2013)Evans and Stanovich]{evans2013dual}
Jonathan St~BT Evans and Keith~E Stanovich.
\newblock Dual-process theories of higher cognition: Advancing the debate.
\newblock \emph{Perspectives on psychological science}, 8\penalty0 (3):\penalty0 223--241, 2013.

\bibitem[Furniturewala et~al.(2024)Furniturewala, Jandial, Java, Banerjee, Shahid, Bhatia, and Jaidka]{furniturewala2024thinking}
Shaz Furniturewala, Surgan Jandial, Abhinav Java, Pragyan Banerjee, Simra Shahid, Sumit Bhatia, and Kokil Jaidka.
\newblock Thinking fair and slow: On the efficacy of structured prompts for debiasing language models.
\newblock \emph{arXiv preprint arXiv:2405.10431}, 2024.

\bibitem[Geva et~al.(2021)Geva, Khashabi, Segal, Khot, Roth, and Berant]{10.1162/tacl_a_00370}
Mor Geva, Daniel Khashabi, Elad Segal, Tushar Khot, Dan Roth, and Jonathan Berant.
\newblock Did aristotle use a laptop? a question answering benchmark with implicit reasoning strategies.
\newblock \emph{Transactions of the Association for Computational Linguistics}, 9:\penalty0 346--361, 04 2021.
\newblock ISSN 2307-387X.
\newblock \doi{10.1162/tacl_a_00370}.
\newblock URL \url{https://doi.org/10.1162/tacl\_a\_00370}.

\bibitem[Gilardi et~al.(2023)Gilardi, Alizadeh, and Kubli]{gilardi2023chatgpt}
Fabrizio Gilardi, Meysam Alizadeh, and Ma{\"e}l Kubli.
\newblock Chatgpt outperforms crowd workers for text-annotation tasks.
\newblock \emph{Proceedings of the National Academy of Sciences}, 120\penalty0 (30):\penalty0 e2305016120, 2023.

\bibitem[Gillan et~al.(2016)Gillan, Kosinski, Whelan, Phelps, and Daw]{gillan2016characterizing}
Claire~M Gillan, Michal Kosinski, Robert Whelan, Elizabeth~A Phelps, and Nathaniel~D Daw.
\newblock Characterizing a psychiatric symptom dimension related to deficits in goal-directed control.
\newblock \emph{elife}, 5:\penalty0 e11305, 2016.

\bibitem[Grootendorst(2022)]{grootendorst2022bertopic}
Maarten Grootendorst.
\newblock Bertopic: Neural topic modeling with a class-based tf-idf procedure.
\newblock \emph{arXiv preprint arXiv:2203.05794}, 2022.

\bibitem[Guo et~al.(2025)Guo, Yang, Zhang, Song, Zhang, Xu, Zhu, Ma, Wang, Bi, et~al.]{guo2025deepseek}
Daya Guo, Dejian Yang, Haowei Zhang, Junxiao Song, Ruoyu Zhang, Runxin Xu, Qihao Zhu, Shirong Ma, Peiyi Wang, Xiao Bi, et~al.
\newblock Deepseek-r1: Incentivizing reasoning capability in llms via reinforcement learning.
\newblock \emph{arXiv preprint arXiv:2501.12948}, 2025.

\bibitem[Hagendorff et~al.(2023)Hagendorff, Fabi, and Kosinski]{hagendorff2023human}
Thilo Hagendorff, Sarah Fabi, and Michal Kosinski.
\newblock Human-like intuitive behavior and reasoning biases emerged in large language models but disappeared in chatgpt.
\newblock \emph{Nature Computational Science}, 3\penalty0 (10):\penalty0 833--838, 2023.

\bibitem[He et~al.(2024)He, Liao, Cao, Liu, Liu, Chen, and Qin]{he-etal-2024-planning}
Tao He, Lizi Liao, Yixin Cao, Yuanxing Liu, Ming Liu, Zerui Chen, and Bing Qin.
\newblock Planning like human: A dual-process framework for dialogue planning.
\newblock In Lun-Wei Ku, Andre Martins, and Vivek Srikumar (eds.), \emph{Proceedings of the 62nd Annual Meeting of the Association for Computational Linguistics (Volume 1: Long Papers)}, pp.\  4768--4791, Bangkok, Thailand, August 2024. Association for Computational Linguistics.
\newblock \doi{10.18653/v1/2024.acl-long.262}.
\newblock URL \url{https://aclanthology.org/2024.acl-long.262/}.

\bibitem[He et~al.(2025)He, Li, Liu, Wang, Bu, Zhang, Peng, Zhang, Zheng, Su, et~al.]{he2025can}
Yancheng He, Shilong Li, Jiaheng Liu, Weixun Wang, Xingyuan Bu, Ge~Zhang, Zhongyuan Peng, Zhaoxiang Zhang, Zhicheng Zheng, Wenbo Su, et~al.
\newblock Can large language models detect errors in long chain-of-thought reasoning?
\newblock \emph{arXiv preprint arXiv:2502.19361}, 2025.

\bibitem[Hosseini et~al.(2014)Hosseini, Hajishirzi, Etzioni, and Kushman]{hosseini-etal-2014-learning}
Mohammad~Javad Hosseini, Hannaneh Hajishirzi, Oren Etzioni, and Nate Kushman.
\newblock Learning to solve arithmetic word problems with verb categorization.
\newblock In Alessandro Moschitti, Bo~Pang, and Walter Daelemans (eds.), \emph{Proceedings of the 2014 Conference on Empirical Methods in Natural Language Processing ({EMNLP})}, pp.\  523--533, Doha, Qatar, October 2014. Association for Computational Linguistics.
\newblock \doi{10.3115/v1/D14-1058}.
\newblock URL \url{https://aclanthology.org/D14-1058/}.

\bibitem[Hu et~al.(2021)Hu, Shen, Wallis, Allen-Zhu, Li, Wang, Wang, and Chen]{hu2021lora}
Edward~J Hu, Yelong Shen, Phillip Wallis, Zeyuan Allen-Zhu, Yuanzhi Li, Shean Wang, Lu~Wang, and Weizhu Chen.
\newblock Lora: Low-rank adaptation of large language models.
\newblock \emph{arXiv preprint arXiv:2106.09685}, 2021.

\bibitem[Hua \& Zhang(2022)Hua and Zhang]{hua-zhang-2022-system}
Wenyue Hua and Yongfeng Zhang.
\newblock System 1 + system 2 = better world: Neural-symbolic chain of logic reasoning.
\newblock In Yoav Goldberg, Zornitsa Kozareva, and Yue Zhang (eds.), \emph{Findings of the Association for Computational Linguistics: EMNLP 2022}, pp.\  601--612, Abu Dhabi, United Arab Emirates, December 2022. Association for Computational Linguistics.
\newblock \doi{10.18653/v1/2022.findings-emnlp.42}.
\newblock URL \url{https://aclanthology.org/2022.findings-emnlp.42/}.

\bibitem[Huang et~al.(2022)Huang, Gu, Hou, Wu, Wang, Yu, and Han]{huang2022large}
Jiaxin Huang, Shixiang~Shane Gu, Le~Hou, Yuexin Wu, Xuezhi Wang, Hongkun Yu, and Jiawei Han.
\newblock Large language models can self-improve.
\newblock \emph{arXiv preprint arXiv:2210.11610}, 2022.

\bibitem[Huang \& Chang(2022)Huang and Chang]{huang2022towards}
Jie Huang and Kevin Chen-Chuan Chang.
\newblock Towards reasoning in large language models: A survey.
\newblock \emph{arXiv preprint arXiv:2212.10403}, 2022.

\bibitem[Huang \& Chang(2023)Huang and Chang]{huang-chang-2023-towards}
Jie Huang and Kevin Chen-Chuan Chang.
\newblock Towards reasoning in large language models: A survey.
\newblock In Anna Rogers, Jordan Boyd-Graber, and Naoaki Okazaki (eds.), \emph{Findings of the Association for Computational Linguistics: ACL 2023}, pp.\  1049--1065, Toronto, Canada, July 2023. Association for Computational Linguistics.
\newblock \doi{10.18653/v1/2023.findings-acl.67}.
\newblock URL \url{https://aclanthology.org/2023.findings-acl.67/}.

\bibitem[Hurst et~al.(2024)Hurst, Lerer, Goucher, Perelman, Ramesh, Clark, Ostrow, Welihinda, Hayes, Radford, et~al.]{hurst2024gpt}
Aaron Hurst, Adam Lerer, Adam~P Goucher, Adam Perelman, Aditya Ramesh, Aidan Clark, AJ~Ostrow, Akila Welihinda, Alan Hayes, Alec Radford, et~al.
\newblock Gpt-4o system card.
\newblock \emph{arXiv preprint arXiv:2410.21276}, 2024.

\bibitem[Jiang et~al.(2023)Jiang, Sablayrolles, Mensch, Bamford, Chaplot, de~las Casas, Bressand, Lengyel, Lample, Saulnier, Lavaud, Lachaux, Stock, Scao, Lavril, Wang, Lacroix, and Sayed]{jiang2023mistral7b}
Albert~Q. Jiang, Alexandre Sablayrolles, Arthur Mensch, Chris Bamford, Devendra~Singh Chaplot, Diego de~las Casas, Florian Bressand, Gianna Lengyel, Guillaume Lample, Lucile Saulnier, Lélio~Renard Lavaud, Marie-Anne Lachaux, Pierre Stock, Teven~Le Scao, Thibaut Lavril, Thomas Wang, Timothée Lacroix, and William~El Sayed.
\newblock Mistral 7b, 2023.
\newblock URL \url{https://arxiv.org/abs/2310.06825}.

\bibitem[Jiang et~al.(2024)Jiang, Xie, Hao, Wang, Mallick, Su, Taylor, and Roth]{jiang-etal-2024-peek}
Bowen Jiang, Yangxinyu Xie, Zhuoqun Hao, Xiaomeng Wang, Tanwi Mallick, Weijie~J Su, Camillo~Jose Taylor, and Dan Roth.
\newblock A peek into token bias: Large language models are not yet genuine reasoners.
\newblock In Yaser Al-Onaizan, Mohit Bansal, and Yun-Nung Chen (eds.), \emph{Proceedings of the 2024 Conference on Empirical Methods in Natural Language Processing}, pp.\  4722--4756, Miami, Florida, USA, November 2024. Association for Computational Linguistics.
\newblock \doi{10.18653/v1/2024.emnlp-main.272}.
\newblock URL \url{https://aclanthology.org/2024.emnlp-main.272/}.

\bibitem[Kahneman(2011)]{kahneman2011thinking}
Daniel Kahneman.
\newblock \emph{Thinking, fast and slow}.
\newblock Farrar, Straus and Giroux, New York, 2011.
\newblock ISBN 9780374275631 0374275637.
\newblock URL \url{https://www.amazon.de/Thinking-Fast-Slow-Daniel-Kahneman/dp/0374275637/ref=wl_it_dp_o_pdT1_nS_nC?ie=UTF8&colid=151193SNGKJT9&coliid=I3OCESLZCVDFL7}.

\bibitem[Kamruzzaman \& Kim(2024)Kamruzzaman and Kim]{kamruzzaman2024prompting}
Mahammed Kamruzzaman and Gene~Louis Kim.
\newblock Prompting techniques for reducing social bias in llms through system 1 and system 2 cognitive processes.
\newblock \emph{arXiv preprint arXiv:2404.17218}, 2024.

\bibitem[Keramati et~al.(2011)Keramati, Dezfouli, and Piray]{keramati2011speed}
Mehdi Keramati, Amir Dezfouli, and Payam Piray.
\newblock Speed/accuracy trade-off between the habitual and the goal-directed processes.
\newblock \emph{PLoS computational biology}, 7\penalty0 (5):\penalty0 e1002055, 2011.

\bibitem[Kojima et~al.(2022)Kojima, Gu, Reid, Matsuo, and Iwasawa]{kojima2022large}
Takeshi Kojima, Shixiang~Shane Gu, Machel Reid, Yutaka Matsuo, and Yusuke Iwasawa.
\newblock Large language models are zero-shot reasoners.
\newblock \emph{Advances in neural information processing systems}, 35:\penalty0 22199--22213, 2022.

\bibitem[Kojima et~al.(2023)Kojima, Gu, Reid, Matsuo, and Iwasawa]{kojima2023largelanguagemodelszeroshot}
Takeshi Kojima, Shixiang~Shane Gu, Machel Reid, Yutaka Matsuo, and Yusuke Iwasawa.
\newblock Large language models are zero-shot reasoners, 2023.
\newblock URL \url{https://arxiv.org/abs/2205.11916}.

\bibitem[Koncel-Kedziorski et~al.(2015)Koncel-Kedziorski, Hajishirzi, Sabharwal, Etzioni, and Ang]{10.1162/tacl_a_00160}
Rik Koncel-Kedziorski, Hannaneh Hajishirzi, Ashish Sabharwal, Oren Etzioni, and Siena~Dumas Ang.
\newblock Parsing algebraic word problems into equations.
\newblock \emph{Transactions of the Association for Computational Linguistics}, 3:\penalty0 585--597, 12 2015.
\newblock ISSN 2307-387X.
\newblock \doi{10.1162/tacl_a_00160}.
\newblock URL \url{https://doi.org/10.1162/tacl\_a\_00160}.

\bibitem[Kong et~al.(2024)Kong, Zhao, Chen, Li, Qin, Sun, Zhou, Wang, and Dong]{kong2024betterzeroshotreasoningroleplay}
Aobo Kong, Shiwan Zhao, Hao Chen, Qicheng Li, Yong Qin, Ruiqi Sun, Xin Zhou, Enzhi Wang, and Xiaohang Dong.
\newblock Better zero-shot reasoning with role-play prompting, 2024.
\newblock URL \url{https://arxiv.org/abs/2308.07702}.

\bibitem[Lakoff(1973)]{lakoff1973hedges}
George Lakoff.
\newblock Hedges: A study in meaning criteria and the logic of fuzzy concepts.
\newblock \emph{Journal of philosophical logic}, 2\penalty0 (4):\penalty0 458--508, 1973.

\bibitem[Lee et~al.(2014)Lee, Shimojo, and O’doherty]{lee2014neural}
Sang~Wan Lee, Shinsuke Shimojo, and John~P O’doherty.
\newblock Neural computations underlying arbitration between model-based and model-free learning.
\newblock \emph{Neuron}, 81\penalty0 (3):\penalty0 687--699, 2014.

\bibitem[Li et~al.(2024)Li, Nourkhiz~Mahjoub, Chalaki, Tadiparthi, Lee, Moradi~Pari, Lewis, and Sycara]{li2024language}
Huao Li, Hossein Nourkhiz~Mahjoub, Behdad Chalaki, Vaishnav Tadiparthi, Kwonjoon Lee, Ehsan Moradi~Pari, Charles Lewis, and Katia Sycara.
\newblock Language grounded multi-agent reinforcement learning with human-interpretable communication.
\newblock \emph{Advances in Neural Information Processing Systems}, 37:\penalty0 87908--87933, 2024.

\bibitem[Ling et~al.(2017)Ling, Yogatama, Dyer, and Blunsom]{ling-etal-2017-program}
Wang Ling, Dani Yogatama, Chris Dyer, and Phil Blunsom.
\newblock Program induction by rationale generation: Learning to solve and explain algebraic word problems.
\newblock In Regina Barzilay and Min-Yen Kan (eds.), \emph{Proceedings of the 55th Annual Meeting of the Association for Computational Linguistics (Volume 1: Long Papers)}, pp.\  158--167, Vancouver, Canada, July 2017. Association for Computational Linguistics.
\newblock \doi{10.18653/v1/P17-1015}.
\newblock URL \url{https://aclanthology.org/P17-1015/}.

\bibitem[Ling et~al.(2025)Ling, Tang, Liu, Yang, Fu, Huang, Huang, Wan, Hou, and Hu]{ling2025wakenllm}
Zipeng Ling, Yuehao Tang, Shuliang Liu, Junqi Yang, Shenghong Fu, Chen Huang, Kejia Huang, Yao Wan, Zhichao Hou, and Xuming Hu.
\newblock Wakenllm: Evaluating reasoning potential and stability in llms via fine-grained benchmarking.
\newblock \emph{arXiv preprint arXiv:2507.16199}, 2025.

\bibitem[Liu et~al.(2022)Liu, Wang, Lin, and Li]{liu2022neural}
Zhixuan Liu, Zihao Wang, Yuan Lin, and Hang Li.
\newblock A neural-symbolic approach to natural language understanding.
\newblock \emph{arXiv preprint arXiv:2203.10557}, 2022.

\bibitem[Magister et~al.(2022)Magister, Mallinson, Adamek, Malmi, and Severyn]{magister2022teaching}
Lucie~Charlotte Magister, Jonathan Mallinson, Jakub Adamek, Eric Malmi, and Aliaksei Severyn.
\newblock Teaching small language models to reason.
\newblock \emph{arXiv preprint arXiv:2212.08410}, 2022.

\bibitem[Mattar \& Daw(2018)Mattar and Daw]{mattar2018prioritized}
Marcelo~G Mattar and Nathaniel~D Daw.
\newblock Prioritized memory access explains planning and hippocampal replay.
\newblock \emph{Nature neuroscience}, 21\penalty0 (11):\penalty0 1609--1617, 2018.

\bibitem[Meng et~al.(2024)Meng, Xia, and Chen]{meng2024simpo}
Yu~Meng, Mengzhou Xia, and Danqi Chen.
\newblock Simpo: Simple preference optimization with a reference-free reward.
\newblock \emph{arXiv preprint arXiv:2405.14734}, 2024.

\bibitem[Miller et~al.(2017)Miller, Botvinick, and Brody]{miller2017dorsal}
Kevin~J Miller, Matthew~M Botvinick, and Carlos~D Brody.
\newblock Dorsal hippocampus contributes to model-based planning.
\newblock \emph{Nature neuroscience}, 20\penalty0 (9):\penalty0 1269--1276, 2017.

\bibitem[Mirzadeh et~al.(2024)Mirzadeh, Alizadeh, Shahrokhi, Tuzel, Bengio, and Farajtabar]{mirzadeh2024gsm}
Iman Mirzadeh, Keivan Alizadeh, Hooman Shahrokhi, Oncel Tuzel, Samy Bengio, and Mehrdad Farajtabar.
\newblock Gsm-symbolic: Understanding the limitations of mathematical reasoning in large language models.
\newblock \emph{arXiv preprint arXiv:2410.05229}, 2024.

\bibitem[Mondorf \& Plank(2024)Mondorf and Plank]{mondorf2024beyond}
Philipp Mondorf and Barbara Plank.
\newblock Beyond accuracy: Evaluating the reasoning behavior of large language models--a survey.
\newblock \emph{arXiv preprint arXiv:2404.01869}, 2024.

\bibitem[Ott(2018)]{ott2018hedging}
Douglas~E Ott.
\newblock Hedging, weasel words, and truthiness in scientific writing.
\newblock \emph{JSLS: Journal of the Society of Laparoendoscopic Surgeons}, 22\penalty0 (4), 2018.

\bibitem[Ouyang et~al.(2022)Ouyang, Wu, Jiang, Almeida, Wainwright, Mishkin, Zhang, Agarwal, Slama, Ray, et~al.]{ouyang2022training}
Long Ouyang, Jeffrey Wu, Xu~Jiang, Diogo Almeida, Carroll Wainwright, Pamela Mishkin, Chong Zhang, Sandhini Agarwal, Katarina Slama, Alex Ray, et~al.
\newblock Training language models to follow instructions with human feedback.
\newblock \emph{Advances in neural information processing systems}, 35:\penalty0 27730--27744, 2022.

\bibitem[Pan et~al.(2024)Pan, Zhang, Zhang, Liu, Wang, and Li]{pan2024dynathink}
Jiabao Pan, Yan Zhang, Chen Zhang, Zuozhu Liu, Hongwei Wang, and Haizhou Li.
\newblock Dynathink: Fast or slow? a dynamic decision-making framework for large language models.
\newblock \emph{arXiv preprint arXiv:2407.01009}, 2024.

\bibitem[Parmar et~al.(2024)Parmar, Patel, Varshney, Nakamura, Luo, Mashetty, Mitra, and Baral]{parmar2024logicbench}
Mihir Parmar, Nisarg Patel, Neeraj Varshney, Mutsumi Nakamura, Man Luo, Santosh Mashetty, Arindam Mitra, and Chitta Baral.
\newblock Logicbench: Towards systematic evaluation of logical reasoning ability of large language models.
\newblock In \emph{Proceedings of the 62nd Annual Meeting of the Association for Computational Linguistics (Volume 1: Long Papers)}, pp.\  13679--13707, 2024.

\bibitem[Patel et~al.(2021)Patel, Bhattamishra, and Goyal]{patel-etal-2021-nlp}
Arkil Patel, Satwik Bhattamishra, and Navin Goyal.
\newblock Are {NLP} models really able to solve simple math word problems?
\newblock In Kristina Toutanova, Anna Rumshisky, Luke Zettlemoyer, Dilek Hakkani-Tur, Iz~Beltagy, Steven Bethard, Ryan Cotterell, Tanmoy Chakraborty, and Yichao Zhou (eds.), \emph{Proceedings of the 2021 Conference of the North American Chapter of the Association for Computational Linguistics: Human Language Technologies}, pp.\  2080--2094, Online, June 2021. Association for Computational Linguistics.
\newblock \doi{10.18653/v1/2021.naacl-main.168}.
\newblock URL \url{https://aclanthology.org/2021.naacl-main.168/}.

\bibitem[Piray \& Daw(2021)Piray and Daw]{piray2021linear}
Payam Piray and Nathaniel~D Daw.
\newblock Linear reinforcement learning in planning, grid fields, and cognitive control.
\newblock \emph{Nature communications}, 12\penalty0 (1):\penalty0 4942, 2021.

\bibitem[Rafailov et~al.(2024)Rafailov, Sharma, Mitchell, Manning, Ermon, and Finn]{rafailov2024direct}
Rafael Rafailov, Archit Sharma, Eric Mitchell, Christopher~D Manning, Stefano Ermon, and Chelsea Finn.
\newblock Direct preference optimization: Your language model is secretly a reward model.
\newblock \emph{Advances in Neural Information Processing Systems}, 36, 2024.

\bibitem[Roy \& Roth(2015)Roy and Roth]{roy-roth-2015-solving}
Subhro Roy and Dan Roth.
\newblock Solving general arithmetic word problems.
\newblock In Llu{\'i}s M{\`a}rquez, Chris Callison-Burch, and Jian Su (eds.), \emph{Proceedings of the 2015 Conference on Empirical Methods in Natural Language Processing}, pp.\  1743--1752, Lisbon, Portugal, September 2015. Association for Computational Linguistics.
\newblock \doi{10.18653/v1/D15-1202}.
\newblock URL \url{https://aclanthology.org/D15-1202/}.

\bibitem[Saha et~al.(2025)Saha, Prasad, Chen, Hase, Stengel-Eskin, and Bansal]{saha2025system}
Swarnadeep Saha, Archiki Prasad, Justin Chen, Peter Hase, Elias Stengel-Eskin, and Mohit Bansal.
\newblock System 1.x: Learning to balance fast and slow planning with language models.
\newblock In \emph{The Thirteenth International Conference on Learning Representations}, 2025.
\newblock URL \url{https://openreview.net/forum?id=zd0iX5xBhA}.

\bibitem[Sap et~al.(2019)Sap, Rashkin, Chen, LeBras, and Choi]{sap2019socialiqa}
Maarten Sap, Hannah Rashkin, Derek Chen, Ronan LeBras, and Yejin Choi.
\newblock Socialiqa: Commonsense reasoning about social interactions.
\newblock \emph{arXiv preprint arXiv:1904.09728}, 2019.

\bibitem[Schad et~al.(2020)Schad, Rapp, Garbusow, Nebe, Sebold, Obst, Sommer, Deserno, Rabovsky, Friedel, et~al.]{schad2020dissociating}
Daniel~J Schad, Michael~A Rapp, Maria Garbusow, Stephan Nebe, Miriam Sebold, Elisabeth Obst, Christian Sommer, Lorenz Deserno, Milena Rabovsky, Eva Friedel, et~al.
\newblock Dissociating neural learning signals in human sign-and goal-trackers.
\newblock \emph{Nature human behaviour}, 4\penalty0 (2):\penalty0 201--214, 2020.

\bibitem[Shojaee et~al.(2025)Shojaee, Mirzadeh, Alizadeh, Horton, Bengio, and Farajtabar]{shojaee2025illusion}
Parshin Shojaee, Iman Mirzadeh, Keivan Alizadeh, Maxwell Horton, Samy Bengio, and Mehrdad Farajtabar.
\newblock The illusion of thinking: Understanding the strengths and limitations of reasoning models via the lens of problem complexity.
\newblock \emph{arXiv preprint arXiv:2506.06941}, 2025.

\bibitem[Singh et~al.(2021)Singh, Wen, Hou, Alipoormolabashi, Wu, Ma, and Peng]{singh-etal-2021-com2sense}
Shikhar Singh, Nuan Wen, Yu~Hou, Pegah Alipoormolabashi, Te-lin Wu, Xuezhe Ma, and Nanyun Peng.
\newblock {COM}2{SENSE}: A commonsense reasoning benchmark with complementary sentences.
\newblock In Chengqing Zong, Fei Xia, Wenjie Li, and Roberto Navigli (eds.), \emph{Findings of the Association for Computational Linguistics: ACL-IJCNLP 2021}, pp.\  883--898, Online, August 2021. Association for Computational Linguistics.
\newblock \doi{10.18653/v1/2021.findings-acl.78}.
\newblock URL \url{https://aclanthology.org/2021.findings-acl.78/}.

\bibitem[Singhal et~al.(2023)Singhal, Goyal, Xu, and Durrett]{Singhal2023ALW}
Prasann Singhal, Tanya Goyal, Jiacheng Xu, and Greg Durrett.
\newblock A long way to go: Investigating length correlations in rlhf.
\newblock \emph{ArXiv}, abs/2310.03716, 2023.
\newblock URL \url{https://api.semanticscholar.org/CorpusID:263672200}.

\bibitem[Sourati et~al.(2024)Sourati, Ilievski, Sommerauer, and Jiang]{sourati2024arn}
Zhivar Sourati, Filip Ilievski, Pia Sommerauer, and Yifan Jiang.
\newblock Arn: Analogical reasoning on narratives.
\newblock \emph{Transactions of the Association for Computational Linguistics}, 12:\penalty0 1063--1086, 2024.

\bibitem[Stanovich \& West(2000)Stanovich and West]{stanovich2000advancing}
Keith~E Stanovich and Richard~F West.
\newblock Advancing the rationality debate.
\newblock \emph{Behavioral and brain sciences}, 23\penalty0 (5):\penalty0 701--717, 2000.

\bibitem[Su et~al.(2025)Su, Sukhbaatar, Rabbat, Tian, and Zheng]{su2025dualformer}
DiJia Su, Sainbayar Sukhbaatar, Michael Rabbat, Yuandong Tian, and Qinqing Zheng.
\newblock Dualformer: Controllable fast and slow thinking by learning with randomized reasoning traces.
\newblock In \emph{The Thirteenth International Conference on Learning Representations}, 2025.
\newblock URL \url{https://openreview.net/forum?id=bmbRCRiNDu}.

\bibitem[Talmor et~al.(2019)Talmor, Herzig, Lourie, and Berant]{talmor-etal-2019-commonsenseqa}
Alon Talmor, Jonathan Herzig, Nicholas Lourie, and Jonathan Berant.
\newblock {C}ommonsense{QA}: A question answering challenge targeting commonsense knowledge.
\newblock In Jill Burstein, Christy Doran, and Thamar Solorio (eds.), \emph{Proceedings of the 2019 Conference of the North {A}merican Chapter of the Association for Computational Linguistics: Human Language Technologies, Volume 1 (Long and Short Papers)}, pp.\  4149--4158, Minneapolis, Minnesota, June 2019. Association for Computational Linguistics.
\newblock \doi{10.18653/v1/N19-1421}.
\newblock URL \url{https://aclanthology.org/N19-1421/}.

\bibitem[Tversky \& Kahneman(1974)Tversky and Kahneman]{tversky1974judgment}
Amos Tversky and Daniel Kahneman.
\newblock Judgment under uncertainty: Heuristics and biases: Biases in judgments reveal some heuristics of thinking under uncertainty.
\newblock \emph{science}, 185\penalty0 (4157):\penalty0 1124--1131, 1974.

\bibitem[Valmeekam et~al.(2022)Valmeekam, Olmo, Sreedharan, and Kambhampati]{valmeekam2022large}
Karthik Valmeekam, Alberto Olmo, Sarath Sreedharan, and Subbarao Kambhampati.
\newblock Large language models still can't plan (a benchmark for llms on planning and reasoning about change).
\newblock In \emph{NeurIPS 2022 Foundation Models for Decision Making Workshop}, 2022.

\bibitem[Wang et~al.(2023)Wang, Xu, Lan, Hu, Lan, Lee, and Lim]{wang-etal-2023-plan}
Lei Wang, Wanyu Xu, Yihuai Lan, Zhiqiang Hu, Yunshi Lan, Roy Ka-Wei Lee, and Ee-Peng Lim.
\newblock Plan-and-solve prompting: Improving zero-shot chain-of-thought reasoning by large language models.
\newblock In Anna Rogers, Jordan Boyd-Graber, and Naoaki Okazaki (eds.), \emph{Proceedings of the 61st Annual Meeting of the Association for Computational Linguistics (Volume 1: Long Papers)}, pp.\  2609--2634, Toronto, Canada, July 2023. Association for Computational Linguistics.
\newblock \doi{10.18653/v1/2023.acl-long.147}.
\newblock URL \url{https://aclanthology.org/2023.acl-long.147/}.

\bibitem[Wang \& Zhou(2024)Wang and Zhou]{wang2024chain}
Xuezhi Wang and Denny Zhou.
\newblock Chain-of-thought reasoning without prompting.
\newblock \emph{arXiv preprint arXiv:2402.10200}, 2024.

\bibitem[Wang et~al.(2022)Wang, Wei, Schuurmans, Le, Chi, Narang, Chowdhery, and Zhou]{wang2022self}
Xuezhi Wang, Jason Wei, Dale Schuurmans, Quoc Le, Ed~Chi, Sharan Narang, Aakanksha Chowdhery, and Denny Zhou.
\newblock Self-consistency improves chain of thought reasoning in language models.
\newblock \emph{arXiv preprint arXiv:2203.11171}, 2022.

\bibitem[Wei et~al.(2022{\natexlab{a}})Wei, Tay, Bommasani, Raffel, Zoph, Borgeaud, Yogatama, Bosma, Zhou, Metzler, et~al.]{wei2022emergent}
Jason Wei, Yi~Tay, Rishi Bommasani, Colin Raffel, Barret Zoph, Sebastian Borgeaud, Dani Yogatama, Maarten Bosma, Denny Zhou, Donald Metzler, et~al.
\newblock Emergent abilities of large language models.
\newblock \emph{arXiv preprint arXiv:2206.07682}, 2022{\natexlab{a}}.

\bibitem[Wei et~al.(2022{\natexlab{b}})Wei, Wang, Schuurmans, Bosma, Xia, Chi, Le, Zhou, et~al.]{wei2022chain}
Jason Wei, Xuezhi Wang, Dale Schuurmans, Maarten Bosma, Fei Xia, Ed~Chi, Quoc~V Le, Denny Zhou, et~al.
\newblock Chain-of-thought prompting elicits reasoning in large language models.
\newblock \emph{Advances in neural information processing systems}, 35:\penalty0 24824--24837, 2022{\natexlab{b}}.

\bibitem[Wei et~al.(2023{\natexlab{a}})Wei, Wang, Schuurmans, Bosma, Ichter, Xia, Chi, Le, and Zhou]{wei2023chainofthoughtpromptingelicitsreasoning}
Jason Wei, Xuezhi Wang, Dale Schuurmans, Maarten Bosma, Brian Ichter, Fei Xia, Ed~Chi, Quoc Le, and Denny Zhou.
\newblock Chain-of-thought prompting elicits reasoning in large language models, 2023{\natexlab{a}}.
\newblock URL \url{https://arxiv.org/abs/2201.11903}.

\bibitem[Wei et~al.(2023{\natexlab{b}})Wei, Cui, Cheng, Wang, Zhang, Huang, Xie, Xu, Chen, Zhang, Jiang, and Han]{wei2023zeroshot}
Xiang Wei, Xingyu Cui, Ning Cheng, Xiaobin Wang, Xin Zhang, Shen Huang, Pengjun Xie, Jinan Xu, Yufeng Chen, Meishan Zhang, Yong Jiang, and Wenjuan Han.
\newblock Zero-shot information extraction via chatting with chatgpt, 2023{\natexlab{b}}.

\bibitem[Weston \& Sukhbaatar(2023)Weston and Sukhbaatar]{weston2023system}
Jason Weston and Sainbayar Sukhbaatar.
\newblock System 2 attention (is something you might need too).
\newblock \emph{arXiv preprint arXiv:2311.11829}, 2023.

\bibitem[Xiao et~al.(2024)Xiao, Xie, Kuang, Liu, Yang, Peng, Han, and Huang]{xiao-etal-2024-healme}
Mengxi Xiao, Qianqian Xie, Ziyan Kuang, Zhicheng Liu, Kailai Yang, Min Peng, Weiguang Han, and Jimin Huang.
\newblock {H}eal{M}e: Harnessing cognitive reframing in large language models for psychotherapy.
\newblock In Lun-Wei Ku, Andre Martins, and Vivek Srikumar (eds.), \emph{Proceedings of the 62nd Annual Meeting of the Association for Computational Linguistics (Volume 1: Long Papers)}, pp.\  1707--1725, Bangkok, Thailand, August 2024. Association for Computational Linguistics.
\newblock \doi{10.18653/v1/2024.acl-long.93}.
\newblock URL \url{https://aclanthology.org/2024.acl-long.93/}.

\bibitem[Xu et~al.(2023)Xu, Sun, Zheng, Geng, Zhao, Feng, Tao, and Jiang]{xu2023wizardlm}
Can Xu, Qingfeng Sun, Kai Zheng, Xiubo Geng, Pu~Zhao, Jiazhan Feng, Chongyang Tao, and Daxin Jiang.
\newblock Wizardlm: Empowering large language models to follow complex instructions.
\newblock \emph{arXiv preprint arXiv:2304.12244}, 2023.

\bibitem[Xu(2021)]{xu2021career}
Hui Xu.
\newblock Career decision-making from a dual-process perspective: Looking back, looking forward.
\newblock \emph{Journal of Vocational Behavior}, 126:\penalty0 103556, 2021.

\bibitem[Yang et~al.(2024)Yang, Shi, Li, Shui, Yang, and Lam]{yang2024llm2}
Cheng Yang, Chufan Shi, Siheng Li, Bo~Shui, Yujiu Yang, and Wai Lam.
\newblock Llm2: Let large language models harness system 2 reasoning.
\newblock \emph{arXiv preprint arXiv:2412.20372}, 2024.

\bibitem[Yao et~al.(2024)Yao, Yu, Zhao, Shafran, Griffiths, Cao, and Narasimhan]{yao2024tree}
Shunyu Yao, Dian Yu, Jeffrey Zhao, Izhak Shafran, Tom Griffiths, Yuan Cao, and Karthik Narasimhan.
\newblock Tree of thoughts: Deliberate problem solving with large language models.
\newblock \emph{Advances in Neural Information Processing Systems}, 36, 2024.

\bibitem[You et~al.(2025)You, Xue, Havaldar, Rao, Jin, Callison-Burch, and Wong]{you2025probabilistic}
Weiqiu You, Anton Xue, Shreya Havaldar, Delip Rao, Helen Jin, Chris Callison-Burch, and Eric Wong.
\newblock Probabilistic soundness guarantees in llm reasoning chains.
\newblock \emph{arXiv preprint arXiv:2507.12948}, 2025.

\bibitem[Yu et~al.(2024)Yu, Xu, Weston, and Kulikov]{yu2024distilling}
Ping Yu, Jing Xu, Jason Weston, and Ilia Kulikov.
\newblock Distilling system 2 into system 1.
\newblock \emph{arXiv preprint arXiv:2407.06023}, 2024.

\bibitem[Zeng et~al.(2024)Zeng, Liu, Wan, Li, Chen, Dai, Yao, Xu, Qi, Zhao, et~al.]{zeng2024mr}
Zhongshen Zeng, Yinhong Liu, Yingjia Wan, Jingyao Li, Pengguang Chen, Jianbo Dai, Yuxuan Yao, Rongwu Xu, Zehan Qi, Wanru Zhao, et~al.
\newblock Mr-ben: A meta-reasoning benchmark for evaluating system-2 thinking in llms.
\newblock In \emph{The Thirty-eighth Annual Conference on Neural Information Processing Systems}, 2024.

\bibitem[Zheng et~al.(2023)Zheng, Chiang, Sheng, Zhuang, Wu, Zhuang, Lin, Li, Li, Xing, et~al.]{zheng2023judging}
Lianmin Zheng, Wei-Lin Chiang, Ying Sheng, Siyuan Zhuang, Zhanghao Wu, Yonghao Zhuang, Zi~Lin, Zhuohan Li, Dacheng Li, Eric Xing, et~al.
\newblock Judging llm-as-a-judge with mt-bench and chatbot arena.
\newblock \emph{Advances in Neural Information Processing Systems}, 36:\penalty0 46595--46623, 2023.

\bibitem[Zhou et~al.(2024{\natexlab{a}})Zhou, Qian, Feng, Hui, Zhu, and Mao]{zhou-etal-2024-llms}
Hanzhang Zhou, Junlang Qian, Zijian Feng, Lu~Hui, Zixiao Zhu, and Kezhi Mao.
\newblock {LLM}s learn task heuristics from demonstrations: A heuristic-driven prompting strategy for document-level event argument extraction.
\newblock In Lun-Wei Ku, Andre Martins, and Vivek Srikumar (eds.), \emph{Proceedings of the 62nd Annual Meeting of the Association for Computational Linguistics (Volume 1: Long Papers)}, pp.\  11972--11990, Bangkok, Thailand, August 2024{\natexlab{a}}. Association for Computational Linguistics.
\newblock \doi{10.18653/v1/2024.acl-long.647}.
\newblock URL \url{https://aclanthology.org/2024.acl-long.647/}.

\bibitem[Zhou et~al.(2024{\natexlab{b}})Zhou, Pujara, Ren, Chen, Cheng, Le, Chi, Zhou, Mishra, and Zheng]{zhou2024self}
Pei Zhou, Jay Pujara, Xiang Ren, Xinyun Chen, Heng-Tze Cheng, Quoc~V Le, Ed~H Chi, Denny Zhou, Swaroop Mishra, and Huaixiu~Steven Zheng.
\newblock Self-discover: Large language models self-compose reasoning structures.
\newblock \emph{arXiv preprint arXiv:2402.03620}, 2024{\natexlab{b}}.

\end{thebibliography}

\appendix

\section{Limitations}\label{appendix:limitation}
Despite the promising advancements of using different thinking styles through the lens of dual-process cognitive theory in our approach, it is important to clarify the intended scope and outline future directions. Our curated dataset of 2,000 questions covers 10 well-established cognitive heuristics and was validated by our domain experts to ensure quality. While not exhaustive, this dataset provides a strong foundation for investigating reasoning style differences and establishes methodological groundwork for broader-scale expansion in future studies to represent the entire spectrum of reasoning challenges encountered in real-world tasks. We focused our alignment experiments on Llama and Mistral as base models, using DPO and SIMPO as preference optimization techniques. While our findings are likely to generalize across model architectures and alignment methods, given the shared emergence of both intuitive and deliberative reasoning in large-scale pretraining, testing this generalization to other architectures and alignment methods is a valuable future direction.
Moreover, while our dynamic model is training-free and improves performance, it is computationally inefficient. It doubles inference costs and memory usage by requiring both models to run for every query. Future work could distill this capability into a single, efficient model to mitigate this overhead.
In terms of evaluating reasoning uncertainty, we adopt token-level logit-based measures and linguistic hedging analysis as computationally tractable proxies. These provide interpretable signals of reasoning behavior, though deeper psycholinguistic and interactive evaluations may offer complementary insights. 
Finally, while our experiments reveal a clear accuracy-efficiency trade-off between intuitive and deliberative reasoning, the extent to which these findings translate to more complex or efficient dynamic decision-making scenarios remains an open question. Future work should explore larger, more diverse datasets and investigate alternative alignment strategies to further validate and extend these results.

\section{Ethical Statement}
Aligning LLMs with $\mathcal{S}1$ and $\mathcal{S}2$ reasoning raises concerns about model behavior in different contexts. On one hand, $\mathcal{S}1$ models risk producing overly confident but incorrect or biased responses, and their alignment with heuristics could be misinterpreted as an endorsement of harmful stereotypes. We want to be clear that the goal of this work is to leverage heuristics for their efficiency, not to amplify unfair biases. On the other hand, $\mathcal{S}2$ models, though more deliberate, are not a universal solution as they introduce slower response times and increased computational costs. Responsible deployment requires building systems that engage the appropriate reasoning style for the context and strike a balance between efficiency and the risk of biased or misleading outputs.

\section{LLM Usage}
We used Large Language Models (specifically OpenAI’s GPT models) exclusively for polishing the writing of this paper. No aspects of the research design, implementation, or analysis involved LLM assistance.

\section{Cognitive heuristics} \label{appendix:cognitivebiases}
In \Cref{tab:cognitive_biases}, we list 10 different cognitive heuristics and their definitions, which we used in curating the dataset \cite{kahneman2011thinking, stanovich2000advancing, evans2013dual}.
\begin{table}[htbp]
    \centering
    \renewcommand{\arraystretch}{1.2}
    \caption{10 common cognitive biases and their definitions, which were considered in curating the dataset}
    \label{tab:cognitive_biases}
    \begin{tabular}{m{3cm}|m{10cm}}
        \toprule
        \textbf{Cognitive Bias} & \textbf{Definition} \\
        \midrule
        Anchoring Bias & The tendency to rely too heavily on the first piece of information we receive about a topic, using it as a reference point for future judgments and decisions, even when new information becomes available. \\
        \midrule
        Halo Effect Bias & The tendency to let one positive impressions of people, brands, and products in one area positively influence our feelings in another area. \\
        \midrule
        Overconfidence Bias & The tendency to have excessive confidence in one's own abilities or knowledge. \\
        \midrule
        Optimism Bias & The tendency to overestimate the likelihood of positive outcomes and underestimate negative ones. \\
        \midrule
        Availability Heuristic Bias & The tendency to use information that comes to mind quickly and easily when making decisions about the future. \\
        \midrule
        Status Quo Bias & The preference for maintaining the current state of affairs, leading to resistance to change.  \\
        \midrule
        Recency Bias & The tendency to better remember and recall information presented to us most recently, compared to information we encountered earlier\\
        \midrule
        Confirmation Bias & The tendency to notice, focus on, and give greater credence to evidence that fits with our existing beliefs.
        \\
        \midrule
        Planning Fallacy & The tendency to underestimate the amount of time it will take to complete a task, as well as the costs and risks associated with that task even if it contradicts our experiences. \\
        \midrule
        Bandwagon Effect Bias & The tendency to adopt beliefs or behaviors because many others do. \\
        \bottomrule
    \end{tabular}
\end{table}

\section{Details of Experts}
\label{appendix:expertsdetails}
The experts consulted are the three authors of this paper: two are PhD students and the other is a faculty member, all specializing in cognitive sciences.

\section{Initial Data Examples} \label{appendix:dataexpertsamples}
\begin{table*}[htbp]
\centering
\tiny
\setlength{\tabcolsep}{4pt}
\resizebox{\textwidth}{!}{
\begin{tabular}{>{\RaggedRight}m{1.5cm} | >{\RaggedRight}m{3cm} | >{\RaggedRight}m{5cm} | >{\RaggedRight}m{5cm}}
\toprule
\multicolumn{1}{c}{\textbf{Category}} & \multicolumn{1}{c}{\textbf{Question}} & \multicolumn{1}{c}{\textbf{System 1 Answer}} & \multicolumn{1}{c}{\textbf{System 2 Answer}} \\
\midrule
Anchoring Bias & 
Do you rely on your first impression of meeting your lab mate ?
& Yes, my gut instinct is usually right.
& I should interact with them more to form a well-rounded opinion.  \\
\midrule
Halo effect Bias & 
How do you feel about the new political candidate?
& I do not like their stance on one issue, so I think they are a terrible candidate.
& I’ll weigh their stance on multiple issues before deciding. \\
\midrule
Over Confidence Bias & 
Do you think you will succeed in your new job?
&
I will definitely succeed here.
& I will need to put in effort and adapt to the new environment to succeed. \\
\midrule
Status Quo Bias & 
Should you change your workout routine?
&
My routine has always worked, so there is no need to change it.
& My fitness needs might have changed, so I will consider adjusting my routine.
\\
\midrule
Optimism Bias & 
Do you need to double-check your work after a mistake?
&
I am usually careful, so one mistake doesn’t mean I’ll make another.
& I will double-check my work to make sure I don’t repeat the mistake. \\
\midrule
Availability heuristic & 
Is the newest seafood restaurant the best restaurant in town?
&
It is the most popular one, so it must be the best.
&
Popularity does not always mean the best quality, so I will read reviews first. 
\\
\midrule
Recency Bias & 
Should you invest in the stock after hearing good things about it?
&
Yes, it is been rising lately, so it’s sure to keep going up.&
I will research the stock and market conditions before making a decision. 
\\
\midrule
Confirmation Bias & 
Is the newest seafood restaurant the best restaurant in town?
&
It is the most popular one, so it must be the best.
&
Popularity does not always mean the best quality, so I will read reviews first. 
\\
\midrule
Planning Fallacy & 
Is the newest seafood restaurant the best restaurant in town?
&
It is the most popular one, so it must be the best.
&
Popularity does not always mean the best quality, so I will read reviews first. 
\\
\midrule
Bandwagon Effect Bias & 
Why did you pick apple as brand of your phone?
&
Everyone I know has this brand, so it must be the best.&
I compared different features and chose the one that suits my needs.
\\
\bottomrule
\end{tabular}
}
\caption{10 samples generated by an expert}
\label{tab:expertsample}
\end{table*}
The 10 samples generated by the expert for our data generation are shown in \Cref{tab:expertsample}.
\section{Prompt for Data Expansion} \label{appendix:dataexpansionprompt}
We expand our sample dataset by concatenating the expert-generated samples with the definitions in \Cref{tab:cognitive_biases}, along with a description of how $\mathcal{S}1$ and $\mathcal{S}2$ would respond to a given question, as shown below:
\begin{tcolorbox}[colback=gray!10, colframe=black, fontupper=\small]
\texttt{The System 1 response should demonstrate intuitive, fast reasoning that relies on the heuristic, showing the shortcut-like process it uses. The responses should highlight the reasoning style itself, not just the
final answer.}
\end{tcolorbox}

\begin{tcolorbox}[colback=gray!10, colframe=black, fontupper=\small]
\texttt{The System 2 response should demonstrate slow, step-by-step reasoning that carefully analyzes the question, explicitly contrasting with System 1. The responses should highlight the reasoning style itself, not just the
final answer.}
\end{tcolorbox}
\section{Final Dataset Sample}
\label{appendix:dataset-sample}

A subset of the curated dataset is shown in \Cref{tab:sampledata}.

\begin{table}[t]
\centering
\caption{Samples of curated dataset with the cognitive heuristic category, each paired with System 1 response using the heuristic and System 2 response mitigating that heuristic.}
\label{tab:sampledata}
\tiny
\setlength{\tabcolsep}{4pt}
\renewcommand{\arraystretch}{0.9}
\begin{tabular}{>{\Centering}m{1cm} | >{\RaggedRight}m{1.5cm} | >{\RaggedRight}m{5cm} | >{\RaggedRight}m{5cm}}
\multicolumn{1}{c}{\textbf{Category}} & \multicolumn{1}{c}{\textbf{Question}} & \multicolumn{1}{c}{\textbf{System 1 Answer}} & \multicolumn{1}{c}{\textbf{System 2 Answer}} \\
\midrule
Anchoring & 
When booking travel, do you find it hard to move away from the first deal you see? 
& I often find myself comparing everything to that initial deal. Once I’ve encountered a specific offer or price, it becomes the standard by which I measure all others. Even when new deals come along, my mind instinctively revisits that first one, guiding my judgment. It’s an automatic comparison that influences how I evaluate options moving forward. 
& To book travel effectively, I avoid comparisons only against the initial option. I understand that the first deal isn't always the best ensures a broader perspective, more observations increases the chances of finding a more favorable travel plan. Exploring more options ultimately leads to better decision-making.  \\
\midrule
Halo Effect  & 
Would you say fasting diet is effective? &
This diet is obviously effective. just look at the people who follow it! Celebrities and influencers swear by it, and they’re in great shape. When so many successful, healthy people use it, that’s proof it works. Plus, it’s super popular, which wouldn’t happen if it didn’t give great results. If you want to see real change, this is the way to go! & 
The effectiveness of a diet depends on scientific evidence, individual health factors, and long-term sustainability rather than its popularity or association with celebrities. A diet may lead to short-term weight loss but could lack essential nutrients or be difficult to maintain. Evaluating its impact on metabolism, overall health, and adherence is crucial.. \\
\midrule
Over Confidence & 
Do you believe you will ace the interview? & 
Yes, I excel in interviews. I communicate clearly, stay confident under pressure, and listen attentively to questions. My ability to understand the interviewer’s needs and align my responses accordingly enhances my effectiveness. I maintain engaging body language and make genuine connections, making a lasting impression. I prepare thoroughly, anticipate potential questions, and rehearse answers, ensuring I approach interviews with a calm, composed demeanor, making me a strong candidate.  & 
While confidence in interpersonal skills is beneficial, thorough interview preparation is essential for success. It allows for anticipation of potential questions and crafting informed responses, showcasing an understanding of the company and role. Researching company culture enables candidates to align their answers with organizational values, enhancing their chances of making a positive impression. Solely relying on confidence can lead to unpreparedness, especially for technical inquiries, reducing the effectiveness of skill articulation.  \\
\bottomrule
\end{tabular}
\end{table}

\section{Topic Modeling} \label{appendix:topicmodeling}
\begin{figure*}[t]
    \centering
    \includegraphics[width=\textwidth]{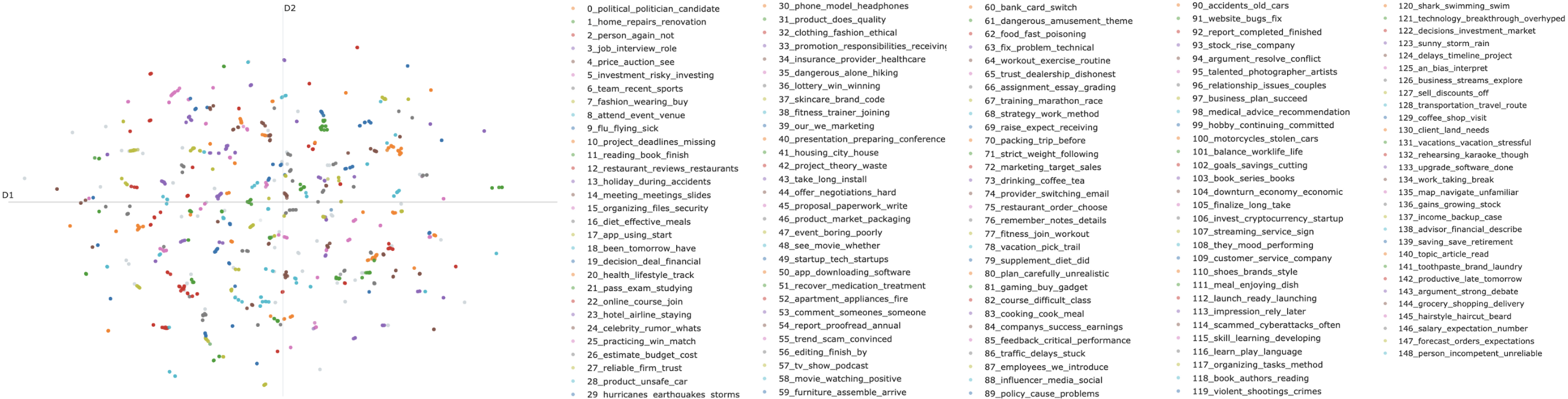}
    \caption{Topic modeling results on our dataset. Each dot represents a question, and colors indicate distinct topics.}
    \label{fig:topic_modeling}
\end{figure*}
Following expert validation, we experimentally verified the diversity of our dataset to ensure it goes beyond surface-level variation in wording. \Cref{fig:topic_modeling}  presents the results of topic modeling using BERTopic \citep{grootendorst2022bertopic}, demonstrating the range of topics covered in the dataset. The wide distribution and clustering across 150 unique topics demonstrate the semantic diversity of the dataset beyond superficial lexical variation.

\section{Benchmark Details} \label{appendix:benchmarkdetails}
We use three categories of reasoning benchmarks: arithmetic, commonsense reasoning, symbolic reasoning,  We provide an overview of the datasets used in each category.
\paragraph{Arithmetic reasoning.}
We use six datasets: MultiArith, GSM8K, AddSub, AQuA, SingleEq, and SVAMP. Each dataset consists of questions that present a scenario requiring numerical computation and multi-step reasoning based on mathematical principles.
\paragraph{Commonsense reasoning.}

To assess commonsense reasoning, we utilize five benchmarks: CommonsenseQA (CSQA), StrategyQA, PIQA, SocialIQA (SIQA), and Com2Sense. All require models to go beyond surface-level understanding and reason using prior knowledge. CSQA focuses on multiple-choice questions grounded in general world knowledge, while StrategyQA includes questions that demand implicit multi-hop reasoning. PIQA evaluates physical commonsense by requiring models to choose the more plausible solution to everyday benchmarks. SIQA targets social commonsense, presenting scenarios about interpersonal interactions and asking questions about motivations, reactions, and emotions. Com2Sense provides pairs of complementary sentences to test a model's ability to distinguish between plausible and implausible statements using commonsense.
\paragraph{Symbolic reasoning.}
We use the Last Letter Concatenation and Coin Flip datasets. Last Letter Concatenation involves forming a word by extracting the last letter of given words in order. Coin Flip presents a sequence of coin-flipping instructions and asks for the final coin orientation. These datasets were originally proposed by \citet{wei2023chainofthoughtpromptingelicitsreasoning} but were not publicly available. \citet{kojima2023largelanguagemodelszeroshot} later followed their approach to create and release accessible versions, which we use in our experiments.

\section{Equivalence Testing of Dataset Lengths Using TOST }\label{appendix:length tost}

A two one-sided t-test (TOST) confirmed the equivalence of these
post-adjustment lengths across various token counts as equivalence margins: $\pm 3$ tokens, $t(3870.30) = 85.82$, $p < .001$; $\pm 5$ tokens, $t(3870.30) = 149.07$, $p < .001$; $\pm 7$ tokens, $t(3870.30) = 212.31$, $p < .001$; and 5\% of the mean token count ($\pm 4.15$ tokens), $t(3870.30) = 122.29$, $p < .001$

\section{Length Adjustment Threshold and Prompt} \label{appendix:length}
We adjust the length if there is a disparity of more than 15 tokens between the $\mathcal{S}1$ and $\mathcal{S}2$ outputs using GPT-4o with the following prompt:

\begin{tcolorbox}[colback=gray!10, colframe=black, fontupper=\small]
\texttt{For a given \{question\}, we have two types of answers:
A fast, intuitive response based on cognitive heuristics which is our System 1 Answer. \\
System 1 Answer: \{System 1 Answer\} \\
And a slow, deliberate, and logical reasoning response which is our System 2 Answer. \\
System 2 Answer: \{System 2 Answer\} \\
Your task is to adjust the two answers so that they are presented in the same order of tokens without altering their content. Ensure that the intuitive nature of the System 1 Answer and the logical reasoning of the System 2 Answer are preserved.}
\end{tcolorbox}

\section{Alignment Algorithm} \label{appendix:alignmentalgorithm}
DPO is an offline alignment method that fine-tunes LLMs by comparing the preferred and disfavored outputs of a model against a reference model, optimizing preferences without requiring a separate reward model. As a prominent method in preference optimization, DPO has gained traction for its stability and efficiency, making it a widely adopted alternative to Reinforcement Learning from Human Feedback \citep[RLHF;][]{ouyang2022training}. SimPO builds on the principles of DPO but introduces a reference-free approach to preference optimization. Instead of requiring a separate reference model, SimPO aligns responses by directly optimizing preference signals within the model itself. This makes it computationally more efficient and removes the dependency on an external reference model, offering a streamlined alternative for aligning LLMs to a specific preference.

\section{Benchmark Instruction} \label{appendix:benchmarkinstruction}
The benchmark-specific instructions are shown in \Cref{tab:benchmarkinstruction}.
\begin{table}[htbp]
    \centering
    \renewcommand{\arraystretch}{1.2} 
    \caption{Benchmark instruction sentences}
    \label{tab:benchmarkinstruction}
    \begin{tabular}{m{4.5cm}|m{4cm}} 
        \toprule
        \textbf{Benchmark} & \textbf{Second Stage Instruction} \\
        \midrule
        MultiArith, SingleEq, AddSub, GSM8K, SVAMP & Therefore, the answer (arabic numerals) is \\
        \midrule
        AQuA, CSQA & Therefore, among A through E, the answer is \\
        \midrule
        SIQA & Therefore, among A through C, the answer is \\
        \midrule
        PIQA & Therefore, among A and B, the answer is \\
        \midrule
        COM2SENSE & Therefore, the answer (TRUE or FALSE) is  \\
        \midrule
        Strategy, Coin & Therefore, the answer (Yes or No) is \\
        \midrule
        Letters & Therefore, the final answer is \\
        \bottomrule
    \end{tabular}
\end{table}

\section{Implementation Details} \label{appendix:impdet}
We use Python 3.10.12, PEFT 0.12.0, PyTorch 2.4.0, and Transformers 4.44.2.
The dataset is split into 80\% training and 20\% validation.
For alignment, we apply Low-Rank Adaptation \citep[LoRA][]{hu2021lora} with a rank of 8, an alpha of 16, and dropout rate of 0.1.
We train for five epochs, using accuracy on winner responses as an early stopping criterion to prevent overfitting, with patience of 5. We set the train batch size to 4 and the validation batch size to 8.
To align Llama 3 using the DPO method, we followed \citet{meng2024simpo} and set the learning rate to $7e-7$ with beta of $0.01$. For SimPO, we use a learning rate of $1e-6$, beta of $2.5$, and a gamma-to-beta ratio of $0.55$. For Mistral v0.1, we set the DPO learning rate to $5e-7$ with beta of $0.001$. In SimPO, we use a learning rate of $5e-7$, beta of $2.5$, and a gamma-to-beta ratio of $0.1$.

The experiments were conducted using NVIDIA RTX A6000 GPU equipped with 48GB of RAM. The total computation time amounted to approximately 800 GPU hours.

\section{Moving from Fast to Slow Thinking Plots}\label{appendix:full-ratio}
\begin{figure*}[t]
\centering
  \includegraphics[width=\columnwidth]{figures/full-ratio_new.png}
  \caption{Accuracy across different benchmarks as reasoning shifts from System 1 to System 2.}
  \label{fig:full-ratio-plot}
\end{figure*}
\Cref{fig:full-ratio-plot}  demonstrates a consistent, monotonic increase in accuracy across all other benchmarks.

\section{Additional Insights into Models' Reasoning}\label{appendix:additional-insights-into-models-reasoning}

In this analysis, we investigate when different models reach definitive answers. We aim to detect this commitment as early as possible during the reasoning process. This early commitment serves as a proxy for the model's confidence in the generated reasoning and its final answer. By analyzing this behavior, we explore whether models can arrive at a definitive answer or if they leave room for ambiguity or subjective interpretation.

We leverage the strong extractive capabilities of LLMs \citep{wei2023zeroshot} and their near-human-like annotation abilities \citep{gilardi2023chatgpt,alizadeh2023open}. Specifically, we focus on the Phi4 (14B) model \citep{abdin2024phi}, which demonstrates exceptional performance in question-answering and reasoning benchmarks, even surpassing closed-source models like GPT-4o \citep{hurst2024gpt}. To determine whether a model’s reasoning contains a definitive answer, we use the following prompt fed to Phi4:

\begin{tcolorbox}[colback=gray!10, colframe=black, fontupper=\small]
    Does the given answer directly answer the given question in a definitive way? ONLY RETURN YES OR NO IN A  \textbackslash textbf\{\}. Definitive answers are clear and do not leave room for interpretation or ambiguity. If the answer tries to explore multiple perspectives or factors involved, it is not definitive, and YOU HAVE TO RETURN NO.
\end{tcolorbox}

\begin{figure*}[htbp]
    \centering
    \includegraphics[width=1\linewidth]{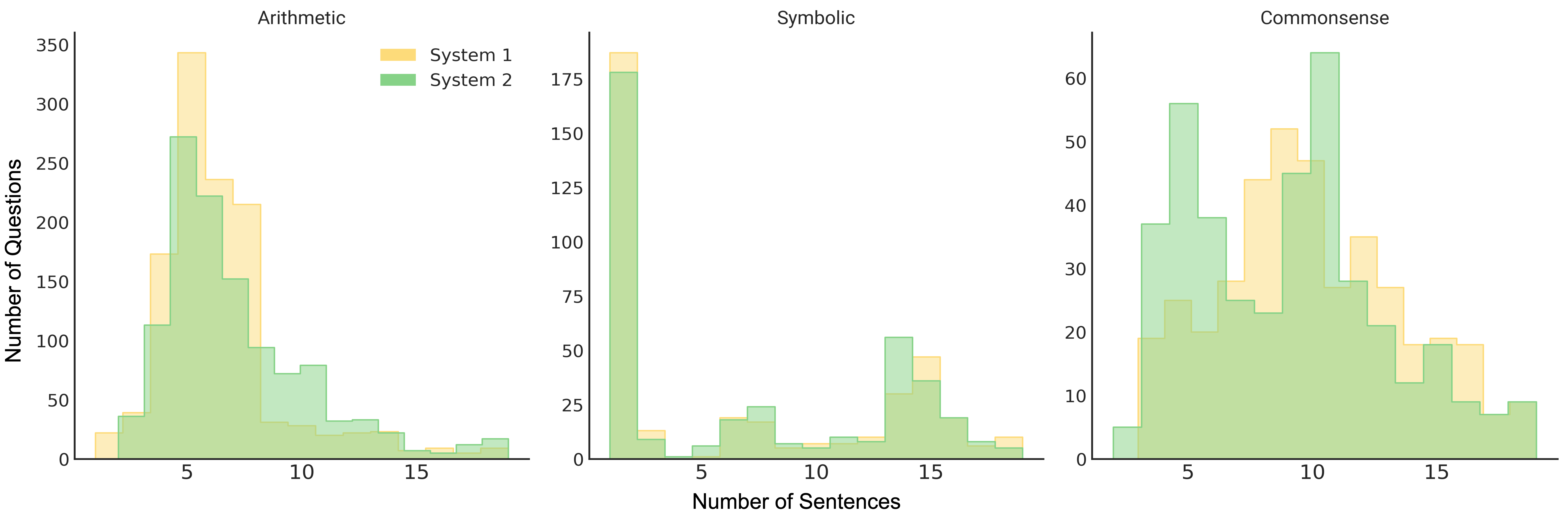}
    \caption{Distribution of the number of sentences in models’ reasoning for both System 1 and System 2 reasoners across different benchmarks.}
    \label{fig:distribution-of-num-sentences-model-reasoning}
\end{figure*}

This prompt is applied to reasoning generated by both $\mathcal{S}1$ and $\mathcal{S}2$ models. To understand when these models commit to a definitive answer during their reasoning process, we focus on the first $n$ sentences of their reasoning, where $n \in \{1, 3, 6, 9, 12, 15\}$. We set a cap of 15 sentences based on our observations that nearly all generated reasonings across benchmarks fall within this range (see \Cref{fig:distribution-of-num-sentences-model-reasoning}).

Applying the prompt to each generated reasoning from the models across all benchmarks (200 randomly sampled data points from each benchmark, totaling 2000 samples for both $\mathcal{S}1$ and $\mathcal{S}2$ reasonings), we append six solved demonstrations to the prompt to help further guide the models. These demonstrations, selected randomly from the cognitive heuristics introduced in \Cref{sec:dataset}, help clarify what qualifies as a definitive answer, aligning the models’ knowledge with patterns we have aligned $\mathcal{S}1$ and 2 models with (see \Cref{subsec:aligning-to-system1-2-thinking}).

\begin{figure*}[htbp]
\centering
  \includegraphics[width=\columnwidth]{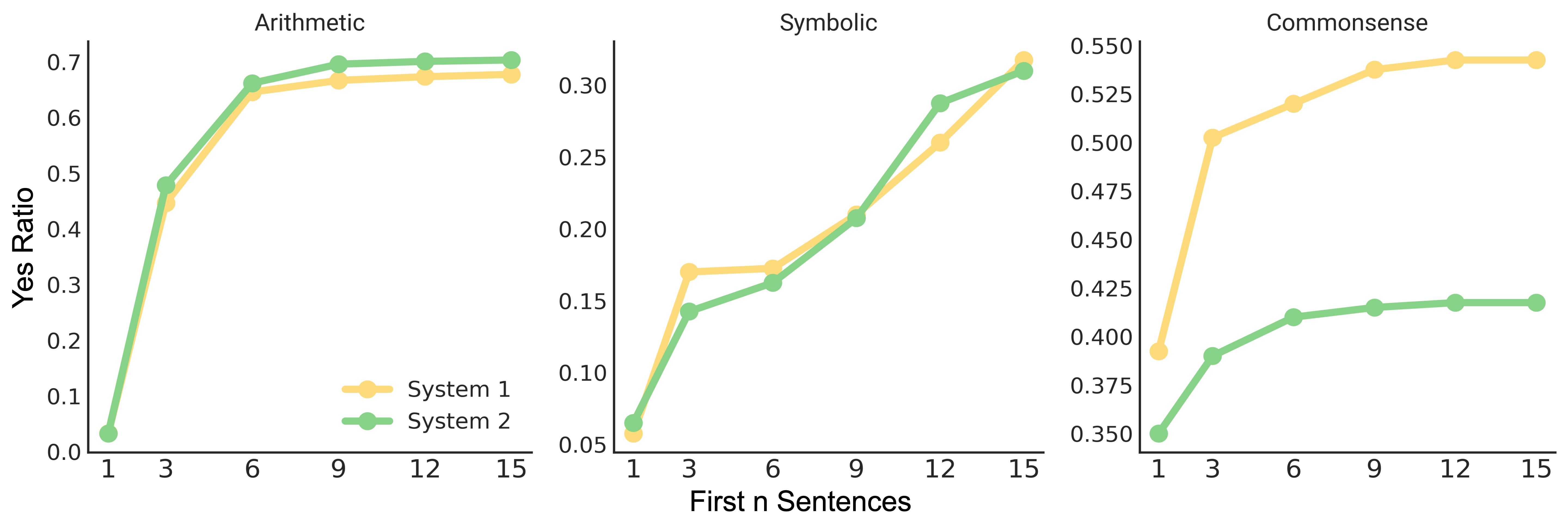}
  \caption{Proportion of definitive answers in the first n sentences across arithmetic, symbolic, and commonsense reasoning tasks}
  \label{fig:exp4_app}
\end{figure*}

\Cref{fig:exp4_app} shows the proportion of definitive answers in the first n sentences, across all benchmarks.\footnote{Note that this ratio should not necessarily converge to 1.0 as more sentences are considered. In some cases, even when considering the full reasoning chain, the models may still leave room for vagueness.} For tasks where quick, intuitive judgments are advantageous, such as in commonsense reasoning. $\mathcal{S}1$ models consistently provide more definitive answers than $\mathcal{S}2$ models. This gap emerges early, with $\mathcal{S}1$ providing more definitive answers in the first three sentences. The difference persists even as we extend the number of sentences considered (see \Cref{tab:mcnemar_results} for a quantitative analysis of the significance between $\mathcal{S}1$ and $\mathcal{S}2$ regarding the definitiveness of their answers).

\begin{table*}[htbp]
\centering
\caption{McNemar's test results comparing the ratio of answers providing committed and definitive responses between System 1 and System 2 across different benchmarks. Statistically significant results ($p$-value < 0.05) are boldfaced.}
\label{tab:mcnemar_results}
\resizebox{\textwidth}{!}{ 
\begin{tabular}{l ccc | ccc | ccc }
    \toprule
    \multirow{2}{*}{\# Sen.} & \multicolumn{3}{c|}{Arithmetic} & \multicolumn{3}{c|}{Symbolic} & \multicolumn{3}{c}{Common Sense} \\
    \cmidrule(lr){2-4} \cmidrule(lr){5-7} \cmidrule(lr){8-10}
    & $\chi^2$ & $p$-value & Winner & $\chi^2$ & $p$-value & Winner & $\chi^2$ & $p$-value & Winner \\
    \midrule
    1  & 21.0  & 1.00  & System 1 & 19.0  & .755  & System 2 & 25.0  & \textbf{.050}  & \textbf{System 1} \\
    3  & 123.0 & \textbf{.028}  & \textbf{System 2} & 29.0  & .228  & System 1 & 20.0  & \textbf{> .001}  & \textbf{System 1} \\
    6  & 125.0 & .272  & System 2 & 33.0  & .720  & System 1 & 21.0  & \textbf{> .001}  & \textbf{System 1} \\
    9  & 120.0 & \textbf{.040}  & \textbf{System 2} & 44.0  & 1.00  & System 1 & 21.0  & \textbf{> .001}  & \textbf{System 1} \\
    12 & 118.0 & .051  & System 2 & 45.0  & .320  & System 2 & 20.0  & \textbf{> .001}  & \textbf{System 1} \\
    15 & 121.0 & .069  & System 2 & 45.0  & .836  & System 1 & 20.0  & \textbf{> .001}  & \textbf{System 1} \\
    \bottomrule
\end{tabular}
}
\end{table*}

\section{System-Specific Failure Patterns}
\label{appendix:failure_patterns}

To complement the main results, we include two analyses that illustrate how $\mathcal{S}1$ and $\mathcal{S}2$ models diverge in failure patterns depending on task type. In numerical reasoning benchmarks, $\mathcal{S}2$ models are more reliable when higher precision is required, while in commonsense benchmarks, $\mathcal{S}1$ models tend to produce more contextually appropriate answers. The following figure and table offer additional insight into these differences.

\begin{figure*}[htbp]
    \centering
    \includegraphics[width=1\linewidth]{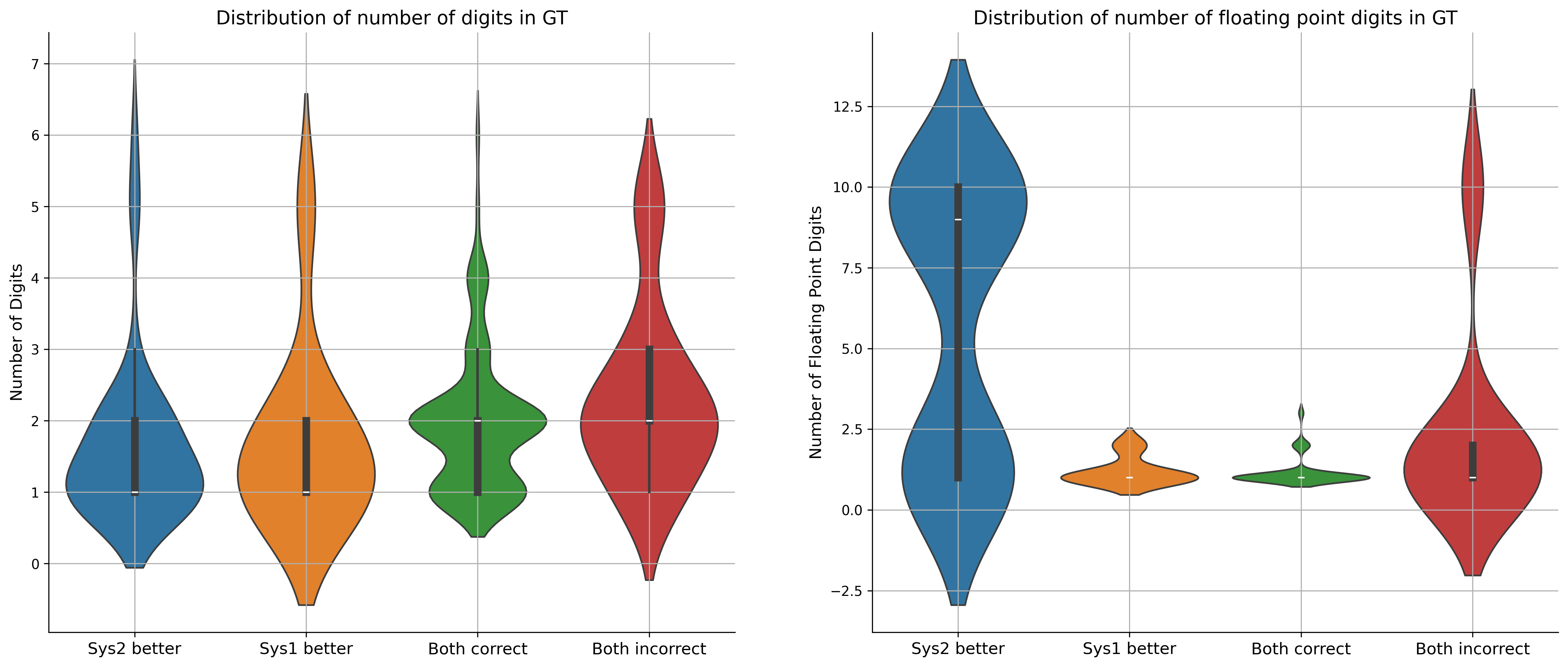}
    \caption{Distribution of digit characteristics in AddSub ground truth (GT) answers across different outcome categories. Left: total number of digits. Right: number of digits after the decimal point. Notably, examples where System 2 is correct and System 1 is not (``Sys2 better'') tend to have more floating-point digits, indicating higher numerical precision.}
    \label{fig:addsub_digit_analysis}
\end{figure*}

To further analyze the behavioral differences between $\mathcal{S}1$ and $\mathcal{S}2$ models, we examine their performance on AddSub items with varying numeric complexity. \Cref{fig:addsub_digit_analysis} shows the distribution of digit types in ground truth answers across four outcome categories. Notably, in examples where $\mathcal{S}2$ succeeds and $\mathcal{S}1$ fails (``Sys2 better''), the ground truth answers tend to have a significantly higher number of floating-point digits (Mann–Whitney U test, $U = 346.0$, $p = 0.0051$). This pattern suggests that $\mathcal{S}2$ is more effective at handling cases requiring greater numerical precision. In contrast, the number of total digits (irrespective of decimal placement) does not differ meaningfully between the ``Sys2 better'' and ``Sys1 better'' subsets ($U = 224.0$, $p = 0.99$).

\vspace{1em}

We also provide a qualitative comparison of commonsense failures made by $\mathcal{S}2$, shown in \autoref{tab:csqa_sys2_failures}. The table includes representative examples from CSQA where $\mathcal{S}2$ responses, although logically coherent, miss intuitive or socially grounded answers. These cases highlight how interpretive depth can lead to answers that diverge from typical human judgment.

\begin{table*}[htbp]
\centering
\tiny
\begin{tabular}{>{\raggedright\arraybackslash}m{4.5cm}|>{\raggedright\arraybackslash}m{4cm}|>{\raggedright\arraybackslash}m{4cm}}
\toprule
\textbf{Question (Answer Choices)} & \textbf{System 2 Rationale \& Final Answer} & \textbf{Analysis and Insight} \\
\midrule
A supermarket is uncommon in what type of collection of shops? (A) strip mall (B) city or town (C) shoppingcentre (D) boutique (E) vermont & Supermarkets are most commonly found in urban or suburban areas such as cities and towns. They are less associated with specific retail arrangements like strip malls or boutiques. \textbf{Final Answer: B} & Overanalyzes spatial patterns instead of recognizing that “strip mall” is a retail form not typically associated with supermarkets. Correct answer: A. \\
\hline
The man tried to reply to the woman, but he had difficulty keeping track of conversations that he didn't do what to? (A) initiate (B) ignore (C) question (D) answer (E) ask & If someone has trouble following conversations, it’s often because they are not actively participating by responding. \textbf{Final Answer: D} & Logical but misses the commonsense interpretation that initiating a conversation anchors attention. Correct answer: A. \\
\hline
What does a kindergarten teacher do before nap time? (A) lower expectations (B) encourage (C) fear (D) time test (E) tell story & Teacher might try to calm children and create a relaxed atmosphere before nap, e.g., by encouraging quiet behavior or mindfulness. \textbf{Final Answer: B} & Misses the stereotypical, culturally familiar act of storytelling before naps. Correct answer: E. \\
\hline
Where can a human find clothes that aren't pants? (A) pants shop (B) on planet earth (C) dress shop (D) school (E) train wreck & On planet Earth, people have access to a wide range of clothing items beyond pants, including dresses, skirts, and robes. \textbf{Final Answer: B} & Too literal and expansive—technically true but ignores the specific, expected answer: C. \\
\hline
Dan was sitting quietly with a book. Laurie thought he was focused, but he actually did what? (A) eat (B) think (C) reading (D) meditate (E) fall asleep & The word “ditting” seems like a typo for “reading,” which matches the context of holding a book. \textbf{Final Answer: C} & Tries to correct perceived error in the question instead of interpreting the pragmatic cue. Correct answer: E. \\
\hline
What instrument can be played with an air of happiness? (A) jump up and down (B) jump up and down (C) sing (D) play games (E) fiddle & Singing is often associated with expressing happiness, and it's a way people naturally convey joy. \textbf{Final Answer: C} & Interprets “instrument” metaphorically rather than literally, overlooking that “fiddle” is an actual musical instrument. Correct answer: E. \\
\hline
If you spend all your time buying and not saving what is likely to happen? (A) using money (B) feel better (C) ocean (D) losing money (E) go broke & Without saving, continuous spending will likely cause one to lose money over time. \textbf{Final Answer: D} & Reasoning is valid, but the phrasing “losing money” is less idiomatic than “go broke,” which better matches the question. Correct answer: E. \\
\hline
Surprising an angry person could lead to what? (A) humor (B) fight (C) jocose (D) laughter (E) accidents & A sudden surprise may defuse tension and even cause laughter, helping to de-escalate anger in some cases. \textbf{Final Answer: D} & Overly optimistic and detached from real-world expectations; more likely outcome is aggression. Correct answer: B. \\
\bottomrule
\end{tabular}
\caption{Examples where the System 2 model fails on commonsense questions. The middle column shows the model’s rationale and final answer, while the right column contains our analysis of why these responses deviate from typical human judgment. In all cases, the System 1 model gives the correct answer.}
\label{tab:csqa_sys2_failures}
\end{table*}

\section{Performance and Validation of the Entropy-Guided Model Selection}
\label{appendix:dynamicappendixres}
\pr{
\Cref{fig:dynamic-plot} presents a comprehensive analysis of our dynamic model, which arbitrates between $\mathcal{S}1$ and $\mathcal{S}2$ responses based on uncertainty signals, as described in \Cref{sec:dynamicmethod}. The line plots for each benchmark show the accuracy for the Llama models, the DPO- and SimPO-aligned dynamic models as weight $w$ from \Cref{eq:combined_score} is varied. The best performance was achieved with $w=0.4$, which supports our hypothesis that penalizing the instability of reasoning more than caution yields more robust results. This demonstrates the effectiveness of our training-free approach in creating a more adaptive reasoning system.
}
\pr{
The analysis of the entropy and variance distributions in the violin plots of \Cref{fig:dynamic-plot} shows a consistent pattern across all benchmarks for the Llama models, the DPO- and SimPO-aligned models, which supports the foundational criteria of our dynamic selection method, as described in \Cref{sec:dynamicmethod}.
}
\pr{
When a system provides a correct answer, its entropy and variance distributions are concentrated in the lower range. This low variance and low entropy case results in the lowest possible score and correctly identifies the response as the most reliable choice. The low variance and high entropy case represents stable but cautious reasoning. Due to the lower weighting of the entropy in our score, this case results in a moderate score, correctly identifying it as a plausible but less confident response. In contrast, incorrect responses are characterized by patterns that lead to higher scores. The high variance and low entropy show the reasoning process is unstable and inconsistent, but the model's average confidence appears high. Our score design addresses this by assigning a greater weight to variance. This ensures that instability is penalized, resulting in a high score that correctly flags the response as unreliable despite its surface-level confidence. The less desirable outcome is the high variance and high entropy case, characterized by reasoning that is both unstable and uncertain. This case results in the highest possible score, correctly identifying it as the least reliable response. Therefore, the systematic separation in these distributions across the four outcome scenarios provides strong empirical evidence that our selection criteria are reliable.
}



\begin{figure*}[htbp]

    \centering
    \includegraphics[width=\linewidth]{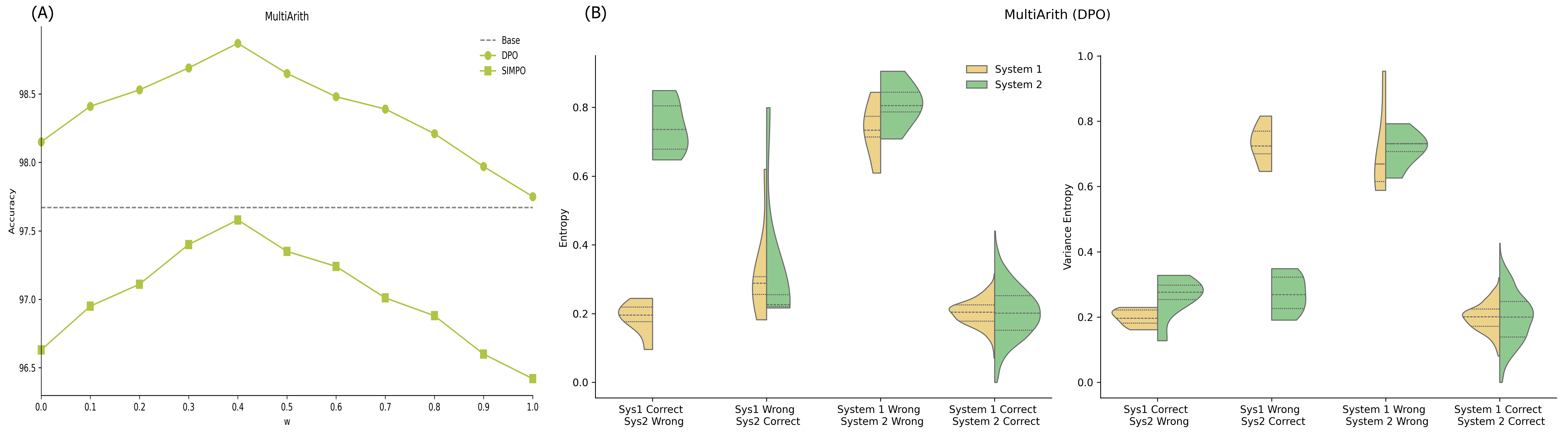}
    \includegraphics[width=\linewidth]{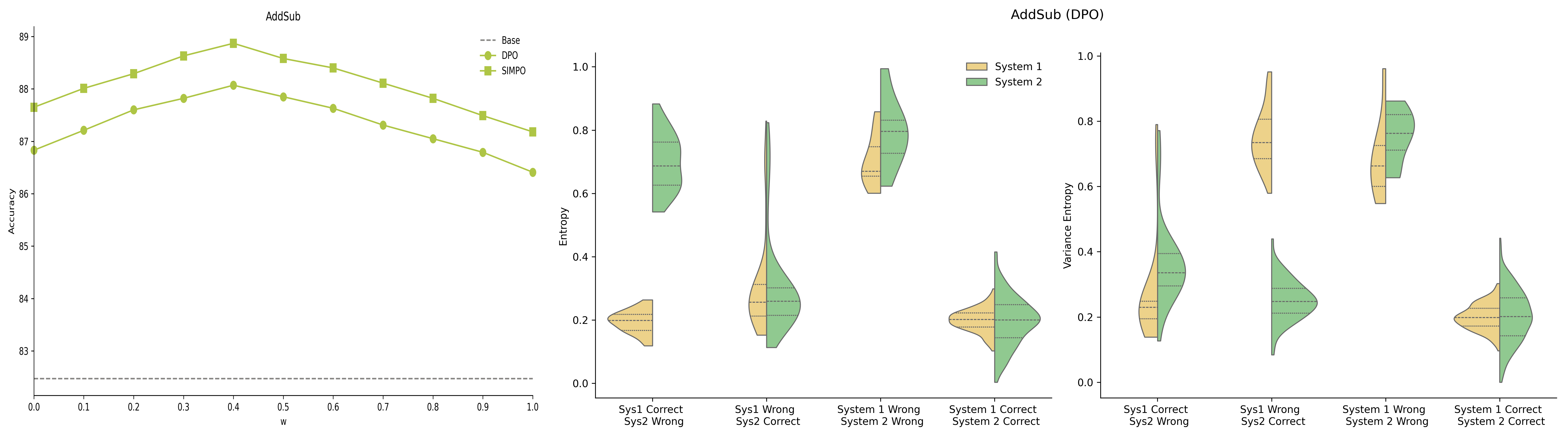}
    \includegraphics[width=\linewidth]{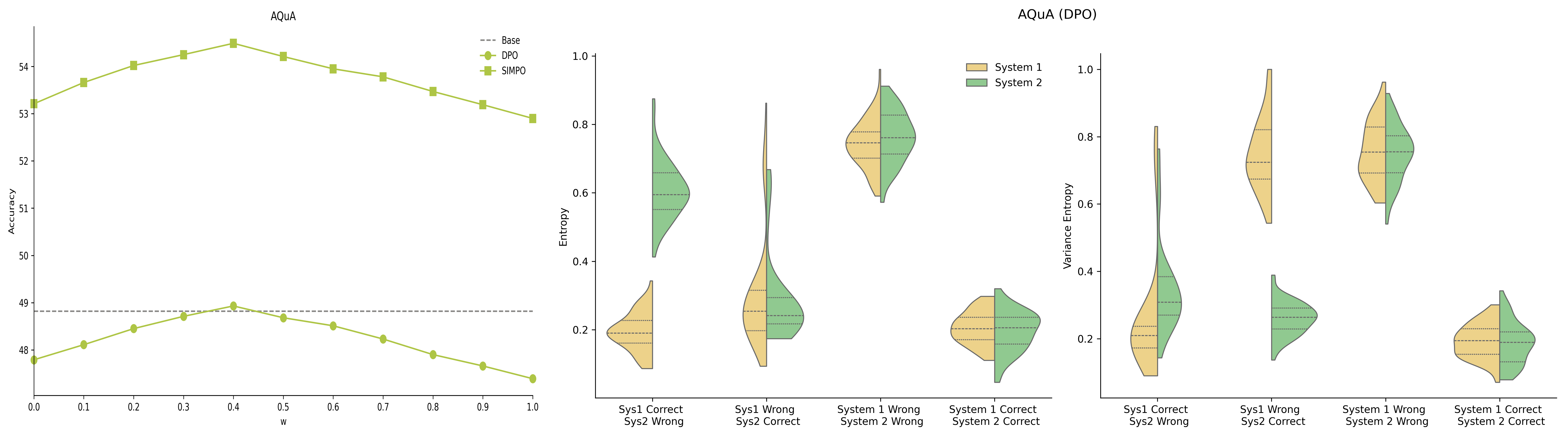}
    \includegraphics[width=\linewidth]{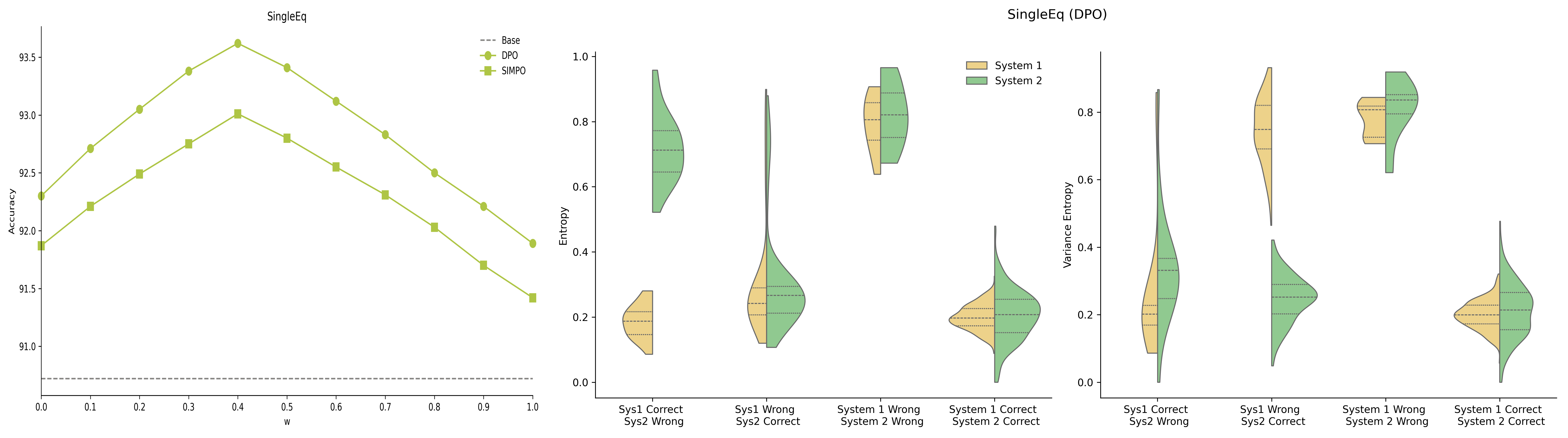}
    \includegraphics[width=\linewidth]{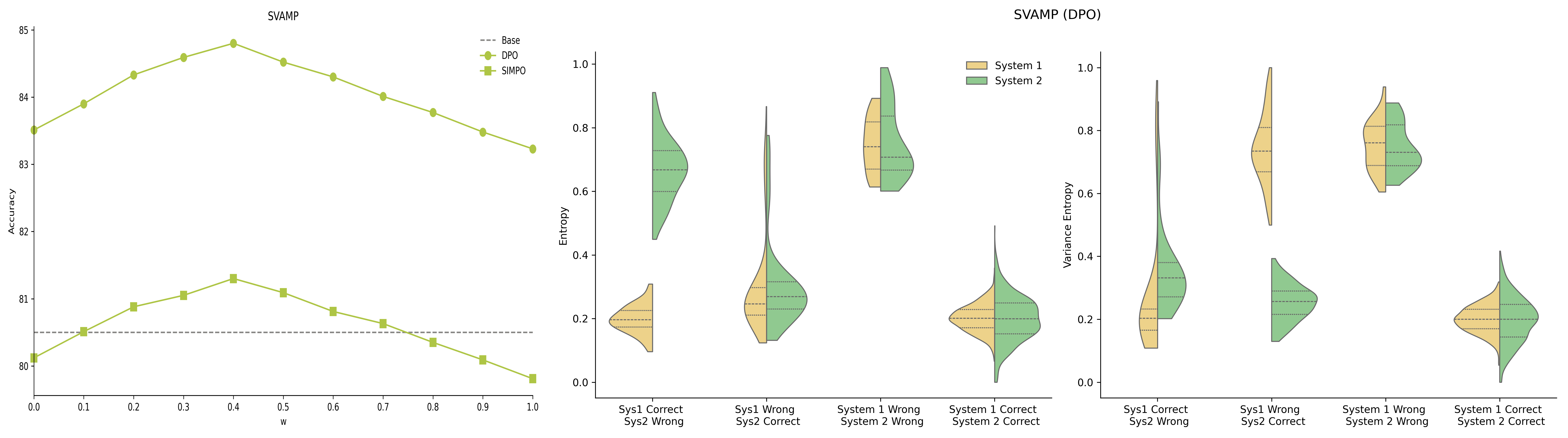}

\caption{}
    \label{fig:dynamic-plot}
\end{figure*}
\begin{figure*}[htbp]
    \ContinuedFloat
    \centering

    \includegraphics[width=\linewidth]{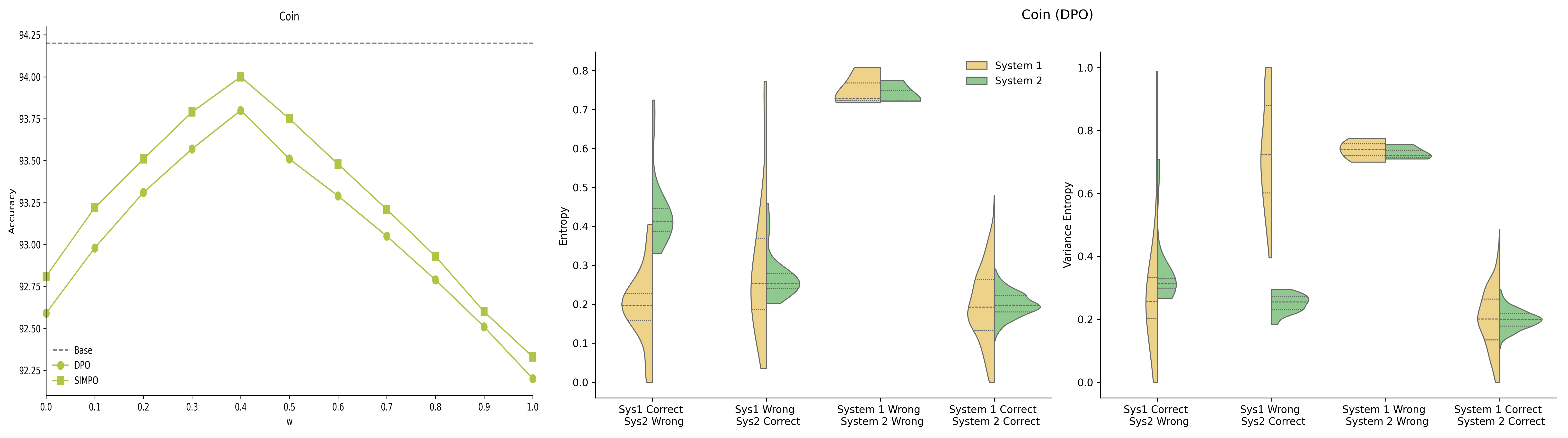}
    \includegraphics[width=\linewidth]{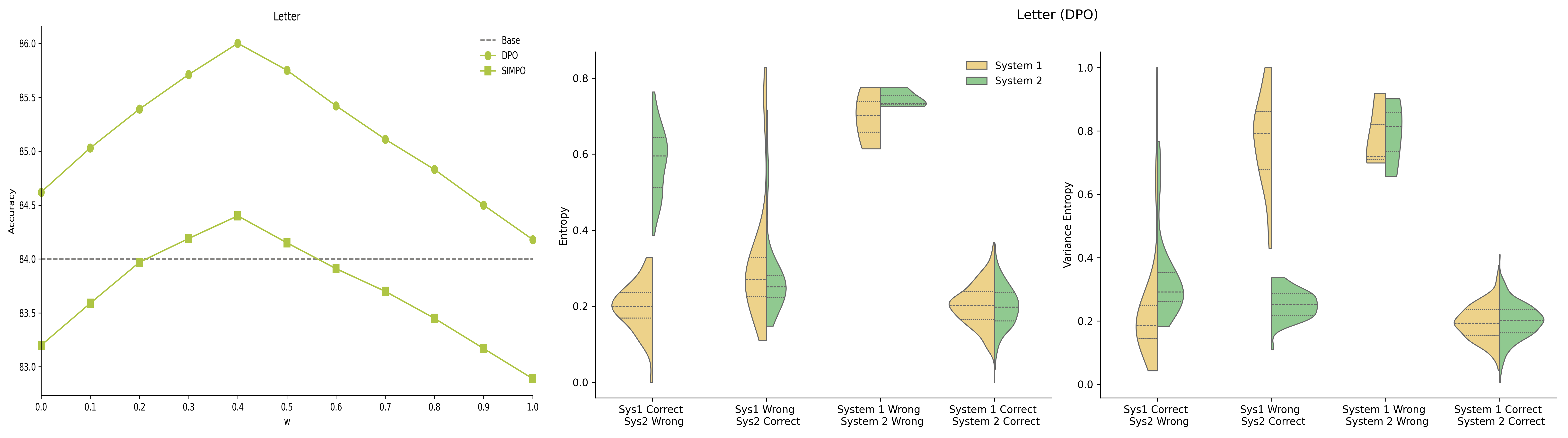}
    \includegraphics[width=\linewidth]{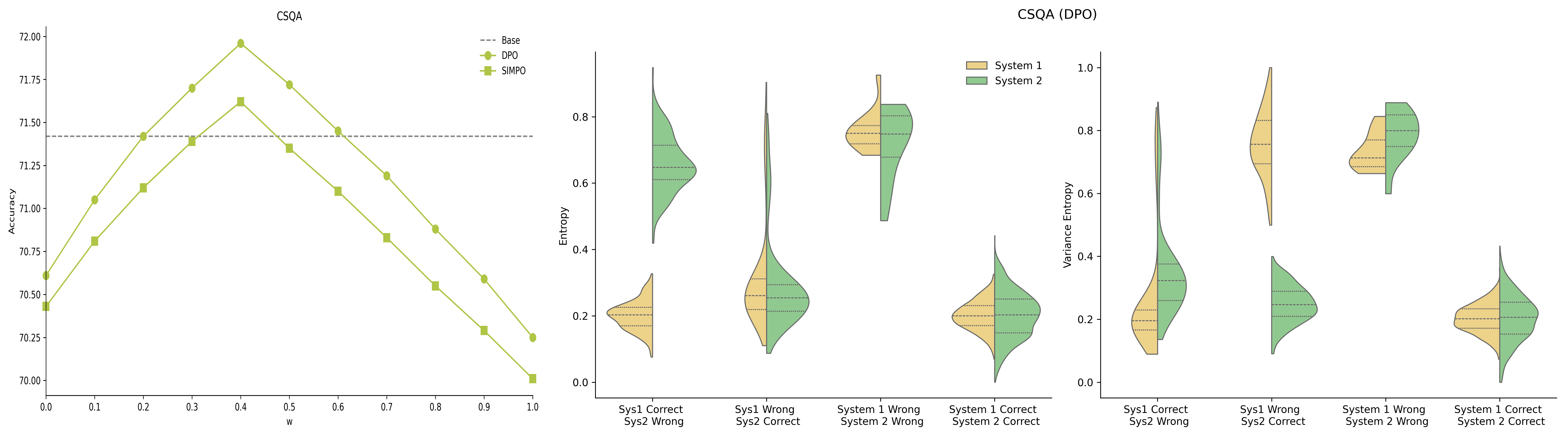}
    \includegraphics[width=\linewidth]{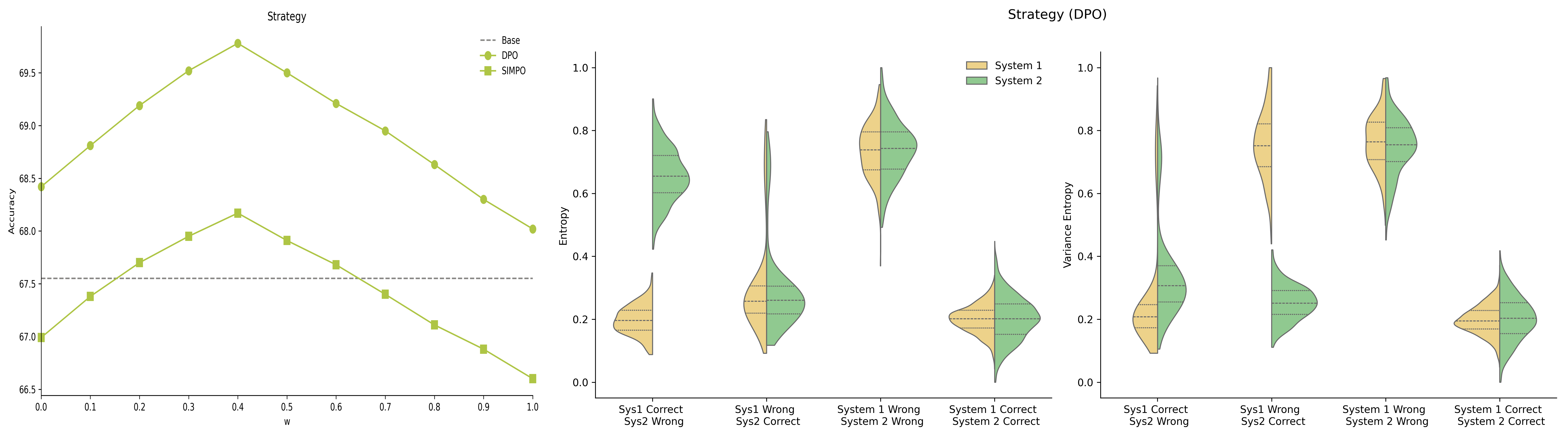}
        \includegraphics[width=\linewidth]{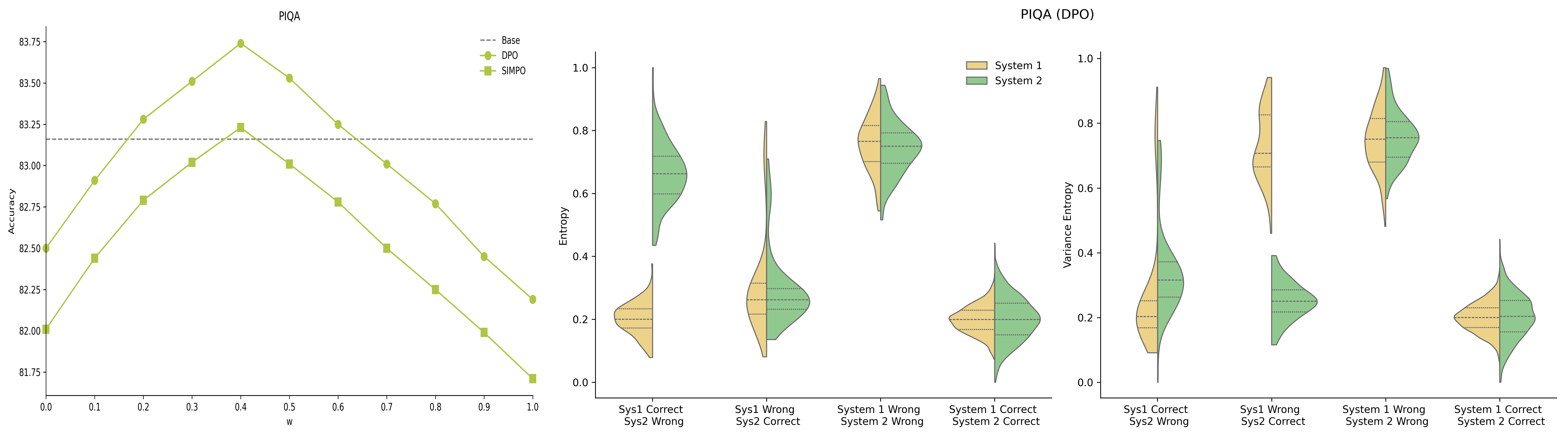}
    \caption[]{}
\end{figure*}

\begin{figure*}[htbp]
    \ContinuedFloat
    \centering

    \includegraphics[width=\linewidth]{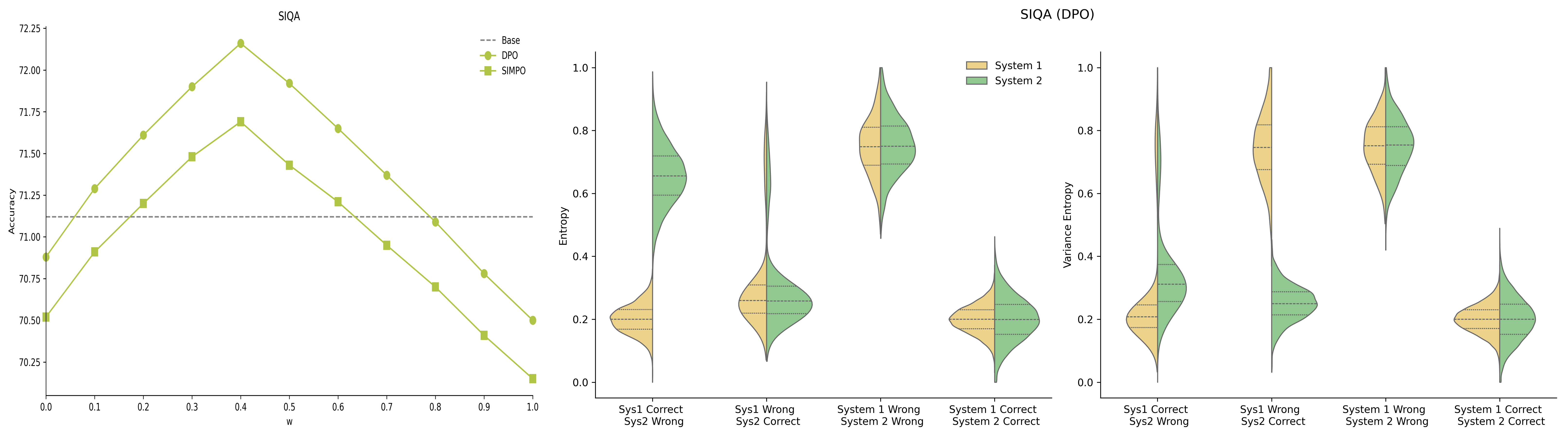}
    \includegraphics[width=\linewidth]{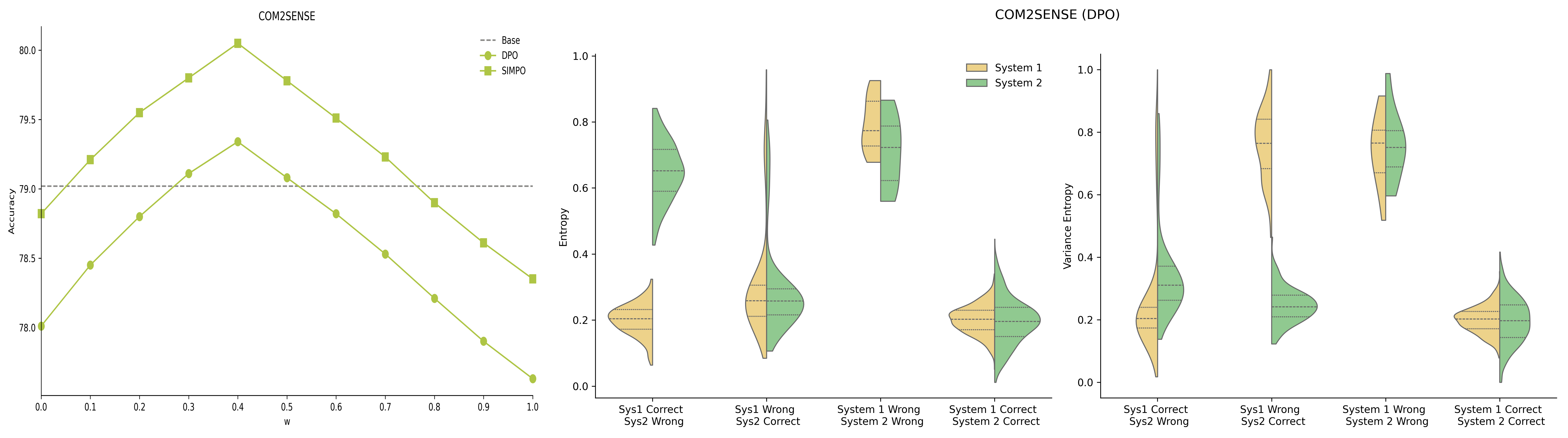}

    \caption[]{Performance of the dynamic model and validation of its entropy-based selection criteria across benchmarks. (A) For each benchmark, the line plot shows the accuracy of the Llama-3 models the DPO- and SimPO-aligned dynamic models as the selection score weight, $w$, is varied. The dashed line represents the accuracy of the base Llama-3 model. (B) The violin plots show the entropy and variance entropy distributions for DPO-aligned Llama models. These distributions are broken down by four distinct outcome scenarios based on the correctness of each system’s response.}
\end{figure*}

\end{document}